\documentclass[11pt]{now} 
\pdfoutput=1

 \usepackage{url}
 \usepackage[numbers]{natbib}
 \usepackage{amsmath,graphicx}

\newcommand{\citeauthornum}[1]{\citeauthor{#1} \cite{#1}}

\newcommand{\by}{\ensuremath{\mathbf{y}}}
\newcommand{\bx}{\ensuremath{\mathbf{x}}}

\newcommand{\y}{\ensuremath{\mathbf{y}}}

\newcommand{\PL}{{\mbox{\tiny \sc pl}}}
\newcommand{\EPL}{{\mbox{\tiny \sc epl}}}

\newcommand{\inst}[2]{{\mathbf #1}^{(#2)}}
\newcommand{\insttvec}[3]{{\mathbf #1}^{(#2)}_{#3}}
\newcommand{\instt}[3]{{#1}^{(#2)}_{#3}}

\newcommand{\lapp}{\ensuremath{\hat{\ell}}}
\newcommand{\lBethe}{\ensuremath{\ell_{\mbox {\tiny \sc Bethe}}}}

\newcommand{\ys}{\ensuremath{\mathbf y}}
\newcommand{\xs}{\ensuremath{\mathbf x}}
\newcommand{\ws}{\ensuremath{\mathbf w}}
\newcommand{\dictV}{{\mathcal V}}

\newcommand{\Data}{{\mathcal D}}
\newcommand{\Ind}[1]{{\mathbf{1}}_{\left\{#1\right\}}}
\newcommand{\defas}{\overset{\mbox{\tiny def}}{=}}

\newcommand{\smCalC}{{\mathcal C}}

\newcommand{\sT}{{\mathrm{\scriptscriptstyle T}}}
\newcommand{\range}[2]{{\langle #1\ldots#2 \rangle}}
\newcommand{\etype}[1]{\textsc{#1}}

\newcommand{\eq}[1]{(\ref{#1})}
\newcommand{\fig}[1]{Figure~\ref{#1}}
\renewcommand{\sec}[1]{Section~\ref{#1}}

\newcommand{\fset}{\mathcal{F}}
\newcommand{\varQFam}{\mathcal{Q}}
\newcommand{\varO}{\mathcal{O}}
\newcommand{\bethe}{{\mbox{\tiny \sc Bethe}}}

\newcommand{\bq}{\mathbf{q}}

\newcommand{\card}[1]{\left\vert #1 \right\vert}

\makeatletter
\newcommand{\Pr@}{\operatorname{Pr}}
\newcommand{\E@}{\operatorname{E}}
\newcommand{\Var@}{\operatorname{Var}}
\renewcommand{\Pr}[1]{\ensuremath{\Pr@\left[{#1}\right]}}
\newcommand{\E}[1]{\ensuremath{\E@\left[{#1}\right]}}
\newcommand{\Var}[1]{\ensuremath{\Var@\left[{#1}\right]}}
\makeatother

\newcommand{\KL}[2]{\ensuremath{\mbox{KL}(#1 \| #2)}}

\newcommand{\MEMM}{\mbox{\tiny MEMM}}

\newtheorem{theorem}{Theorem}[chapter]
\newtheorem{definition}{Definition}[chapter]

\begin{document}

\thispagestyle{empty}

\begin{center}
{\fontsize{16}{32}\selectfont An Introduction to Conditional Random Fields}

\vspace{20pt}

{Charles Sutton\\ University of Edinburgh\\ \url{csutton@inf.ed.ac.uk} \\
 \vspace{1em}\ \\ Andrew McCallum \\ University of Massachusetts Amherst \\ \url{mccallum@cs.umass.edu}}
\vspace{20pt}

{17 November 2010}

\vspace{30pt}

\noindent {\large \bf Abstract}
\end{center}

\noindent
Often we wish to predict a large
  number of variables that depend on each other as well as on other
  observed variables.  Structured
  prediction methods are essentially a combination of classification
  and graphical modeling, combining the ability of graphical models to
  compactly model multivariate data with the ability of classification
  methods to perform prediction using large sets of input features.
  This tutorial describes \emph{conditional random fields}, a
  popular probabilistic method for structured prediction.  
  CRFs have seen wide application in natural language processing,
  computer vision, and bioinformatics.
  We describe methods for inference and parameter estimation for
  CRFs, including practical issues for implementing large
  scale CRFs.  We do not assume previous knowledge of graphical
  modeling, so this tutorial is intended to be useful to practitioners
  in a wide variety of fields.

\vfil


\clearpage \pagenumbering{roman}

\tableofcontents

\clearpage

\setcounter{page}{0}
\pagenumbering{arabic}

\chapter{Introduction}

Fundamental to many applications is the ability to predict multiple
variables that depend on each other.  Such applications are as diverse
as classifying regions of an image \cite{li01mrf}, estimating the
score in a game of Go \cite{stern05go}, segmenting genes in a strand
of DNA \cite{bernal07gene}, and extracting syntax from
natural-language text \cite{taskar04mmcfg}.  In such applications, we
wish to predict a vector $\by=\{y_{0}, y_{1}, \ldots, y_{T} \}$ of
random variables given an observed feature vector $\bx$.  A relatively
simple example from natural-language processing is part-of-speech
tagging, in which each variable $y_{s}$ is the part-of-speech tag of the word
at position $s$, and the input $\bx$ is divided into feature vectors
$\{ \bx_{0}, \bx_{1} \ldots \bx_{T} \}$.  Each $\bx_{s}$ contains
various information about the word at position $s$, such as its
identity, orthographic features such as prefixes and suffixes,
membership in domain-specific lexicons, and information in semantic
databases such as WordNet.

One approach to this multivariate prediction problem, especially if our goal is to
maximize the number of labels $y_{s}$ that are correctly classified, is to
learn an independent per-position classifier that maps $\bx \mapsto y_{s}$ for each
$s$.  The difficulty, however, is that the output variables have complex dependencies.
For example, neighboring words in a document or neighboring regions in a image tend to have similar labels.
Or the output variables may represent a complex structure such as a parse tree,
in which a choice of what grammar rule to use near the top of the tree can have a large effect on the rest of the tree.

A natural way to represent the manner in which output variables depend on
each other is provided by graphical models.  Graphical models---which
include such diverse model families as Bayesian networks, neural
networks, factor graphs, Markov random fields, Ising models, and
others---represent a complex distribution over many
variables as a product of local \emph{factors} on smaller
subsets of variables.  
It is then possible to describe
how a given {factorization} of the
probability density corresponds to a particular set of {conditional
  independence} relationships satisfied by the distribution.  This correspondence
makes modeling much more convenient, because often our knowledge of
the domain suggests reasonable conditional independence assumptions,
which then determine our choice of factors.

Much work in learning with graphical models, especially
 in statistical natural-language processing, has focused
on \emph{generative} models that explicitly attempt to model a
joint probability distribution $p(\by,\bx)$ over inputs and outputs.  
Although there are advantages to this approach, it also has
important limitations. 
Not only can the dimensionality of $\bx$ be very large, but the features
have complex dependencies, so constructing a probability distribution over them
is difficult.  Modelling the dependencies among inputs can lead to intractable models,
but ignoring them 
can lead to reduced performance.

A solution to this problem is to model the conditional distribution $p(\by | \bx)$
directly, which is all that is needed for classification.
This is a conditional random field (CRF).
CRFs are essentially a way of combining the advantages of classification and
graphical modeling, combining the ability to compactly model
multivariate data with the ability to leverage a large number of input
features for prediction.  
The advantage to a conditional model is that dependencies that involve only
variables in $\bx$ play no role in the conditional model, so that an accurate
conditional model can have much simpler structure than a joint model.
The difference between generative models and CRFs is thus exactly analogous
to the difference between the naive Bayes and logistic regression
classifiers.  Indeed, the multinomial logistic regression model can
be seen as the simplest kind of CRF, in which there is only one output
variable.  

There has been a large amount of applied interest in CRFs.  Successful
applications have included text processing
\citep{peng04ie,settles05abner,sha03shallow},
bioinformatics \citep{sato05rna,liu06segmentation}, and computer
vision \citep{he04mcrfs,kumar03drf}. 
Although 
early applications of CRFs used linear chains, recent
applications of CRFs have also used more general graphical
structures. 
General graphical structures 
are useful for predicting complex structures, such as
graphs and trees, and for 
relaxing the iid assumption among entities,
as in relational learning \cite{taskar02rmn}.

This tutorial describes modeling, inference, and parameter estimation
using conditional random fields.
We do not assume previous knowledge of graphical
  modeling, so this tutorial is intended to be useful to practitioners
  in a wide variety of fields.
  We begin by describing modelling
issues in CRFs (Chapter~\ref{chp:graphical}), including linear-chain
CRFs, CRFs with general graphical structure, and hidden CRFs that
include latent variables.  We describe how CRFs can be viewed both as
a generalization of the well-known logistic regression procedure, and
as a discriminative analogue of the hidden Markov model. 

 In the next
two chapters, we describe inference (Chapter~\ref{chp:inference}) and
learning (Chapter~\ref{chp:training}) in CRFs.  The two procedures are
closely coupled, because learning usually calls inference as a
subroutine.  Although the inference algorithms that we discuss are
standard algorithms for graphical models, the fact that inference is
embedded within an outer parameter estimation procedure raises
additional issues.
Finally, we discuss relationships between CRFs and other families of
models, including other structured prediction methods, neural
networks, and maximum entropy Markov models (Chapter~\ref{chp:related}).

\section*{Implementation Details}

Throughout this monograph, we try to point out implementation details
that are sometimes elided in the research literature.  For example, we
discuss issues relating to feature engineering
(Section~\ref{sec:feature-engineering}), avoiding numerical
overflow during inference (Section~\ref{inf:sec:implementation}),
and the scalability of CRF training on some benchmark problems (Section~\ref{sec:times}).

Since this is the first of our sections on implementation details, it
seems appropriate to mention some of the available implementations of CRFs.
At the time of writing, a few popular implementations are:

\begin{table}[h!]
  \centering
  \begin{tabular}{l|l}
\hline
CRF++ & \url{http://crfpp.sourceforge.net/} \\
MALLET & \url{http://mallet.cs.umass.edu/} \\
GRMM & \url{http://mallet.cs.umass.edu/grmm/} \\
CRFSuite & \url{http://www.chokkan.org/software/crfsuite/} \\
FACTORIE & \url{http://www.factorie.cc} \\
\hline
\end{tabular}
\end{table}

Also, software for Markov Logic networks (such as Alchemy:
\url{http://alchemy.cs.washington.edu/}) can be used to build CRF
models.  Alchemy, GRMM, and FACTORIE are the only toolkits of which we are aware
that handle arbitrary graphical structure.

\chapter{Modeling}
\label{chp:graphical} 

In this chapter, we describe conditional random fields from a modeling
perspective, explaining how a CRF represents distributions over
structured outputs as a function of a high-dimensional input vector.
CRFs can be understood both
as an extension of the logistic regression classifier to arbitrary
graphical structures, or as a discriminative analog of generative
models of structured data, an such as hidden Markov models.

We begin with a brief introduction to graphical modeling
(Section~\ref{sec:graphical}) and a description of generative and
discriminative models in NLP (Section~\ref{sec:gendisc}).
Then we will be able to present the formal definition of conditional
random field, both for the commonly-used case of linear chains
(Section~\ref{sec:lc-crf-defn}), and for general graphical structures
(Section~\ref{sec:mdl:crf-general}).  Finally, we present some
examples of how different structures are used in applications
(Section~\ref{chp:applications}), and some implementation details
concerning feature engineering (Section~\ref{sec:feature-engineering}).

\section{Graphical Modeling}
\label{sec:graphical}

Graphical modeling is a powerful framework for representation and 
inference in multivariate probability distributions.  It has proven 
useful in diverse areas of stochastic modeling, including coding 
theory \citep{mceliece98turbo}, computer vision \citep{geman84gibbs}, knowledge 
representation \citep{pearl88priis}, Bayesian statistics \citep{gelfand90sampling}, 
and natural-language processing \citep{lafferty01crf,blei03latent}. 
 
Distributions over many 
variables can be expensive to represent na\"ively.  For example, 
a table of joint probabilities of $n$ binary variables requires storing $O(2^n)$ floating-point numbers. 
The insight of 
the graphical modeling perspective is that a distribution 
over very many variables can often be represented as 
a product of \emph{local functions} that each depend on a much smaller 
subset of variables.  This factorization turns out to 
have a close connection to certain conditional independence relationships among 
the variables---both types of information being easily summarized by a graph. 
Indeed, this relationship between 
factorization, conditional independence, and graph structure comprises much of the 
power of the graphical modeling framework: the conditional 
independence viewpoint is most useful for designing models, and the factorization 
viewpoint is most useful for designing inference algorithms. 
 
In the rest of this section, we introduce graphical models from both 
the factorization and conditional independence viewpoints, focusing on 
those models which are based on undirected graphs. 
A more detailed modern perspective on graphical modelling and
approximate inference is available in a textbook by \citeauthornum{koller:book}.
 
\subsection{Undirected Models} 
 
We consider probability distributions over sets of random variables $V 
= X \cup Y$, where $X$ is a set of \emph{input variables} that we 
assume are observed, and $Y$ is a set of \emph{output variables} that 
we wish to predict.  Every variable $s \in V$ takes outcomes from a 
set $\dictV$, which can be either continuous or discrete, although we
consider only the discrete case in this tutorial.  An arbitrary 
assignment to $X$ is denoted by a vector $\xs$.  Given a variable $s 
\in X$, the notation $x_s$ denotes the value assigned to $s$ by $\bx$, 
and similarly for an assignment to a subset $a \subset X$ by $\xs_a$. 
The notation $\Ind{x=x'}$ denotes an indicator function of $x$ which 
takes the value 1 when $x=x'$ and 0 otherwise.  We also require 
notation for marginalization.  For a fixed variable assignment 
$y_s$, we use the summation $\sum_{\ys \backslash y_s}$ to indicate a 
summation over all possible assignments $\ys$ whose value for variable 
$s$ is equal to $y_s$. 
 
  Suppose that we  believe that a probability 
  distribution $p$ of interest can be represented by a product of 
  \emph{factors} of the form $\Psi_a (\xs_a, \ys_a)$, where each factor has scope $a 
  \subseteq V$.  This factorization can allow us to represent $p$ much 
  more efficiently, because the sets $a$ may be much smaller than the 
  full variable set $V$.  We assume that without loss of generality that each distinct set $a$ has at most one factor $\Psi_a$.
 
An undirected graphical model is a family of probability distributions that factorize according to given collection of scopes. 
Formally, given a collection of subsets $\fset = a \subset V$, an \emph{undirected graphical model} is defined 
as the set of all distributions that can be written in the form 
\begin{equation} 
p(\xs, \ys) = \frac{1}{Z} \prod_{a \in \fset} \Psi_a (\xs_a, \ys_a)\label{sm:eqn:factorization}, 
\end{equation} 
for any choice of \emph{local function} $F = \{ \Psi_a \}$, where $\Psi_a : \dictV^{\card{a}} 
 \rightarrow \Re^+$.  (These functions are also called 
 \emph{factors} or \emph{compatibility functions}.)  We will occasionally use the term \emph{random field} to refer to a particular 
distribution among those defined by an undirected model.   
The reason for the term \emph{graphical model} will become apparent shortly, when we discuss how the factorization of \eq{sm:eqn:factorization} can be represented as a graph. 
 
The constant $Z$ is a normalization factor that ensures the distribution $p$ sums to 1.  It is defined as 
\begin{equation} 
Z = \sum_{\xs, \ys}  \prod_{a \in \fset} \Psi_a (\xs_a, \ys_a).\label{sm:eqn:local-expfam} 
\end{equation} 
The quantity $Z$, 
considered as a function of the set $F$ of factors, 
is sometime called the \emph{partition function}.  Notice that the
summation in \eqref{sm:eqn:local-expfam} is over the exponentially many possible
assignments to $\bx$ and $\by$.  For this reason, computing $Z$ is intractable in general, 
but much work exists on how to approximate it.

We will generally assume further that each local function has the form 
\begin{equation} 
\Psi_a (\xs_a, \ys_a) = \exp \left\{ \sum_{k} \theta_{ak} f_{ak} (\xs_a, \ys_a) \right\}, 
\label{sm:eqn:exponential-factor} 
\end{equation} 
for some real-valued parameter vector $\theta_a$, and for some set of 
\emph{feature functions} or \emph{sufficient statistics} $\{ f_{ak} 
\}$. If $\bx$ and $\by$ are discrete, then this assumption is without loss of generality,
because we can have features have indicator functions for every possible value, 
that is, if we include one feature function $f_{ak} (\xs_a, \ys_a) = \Ind{\xs_a = \xs_a^*} \Ind{\ys_a = \ys_a^*}$  
for every possible value $\xs_a^*$ and $\ys_a^*$. 
 
Also, a consequence of this parameterization is that the family of 
distributions over $V$ parameterized by $\theta$ is an exponential 
family.  Indeed, much of the discussion in this tutorial about
parameter estimation for CRFs applies to exponential families in general. 
 
As we have mentioned, there is a close connection between the 
factorization of a graphical model and the conditional independencies 
among the variables in its domain.  This connection can be understood 
by means of an undirected graph known as a \emph{Markov 
  network}, which directly represents 
conditional independence relationships in a multivariate distribution. 
Let $G$ be an undirected graph with variables $V$, that is, $G$ has 
one node for every random variable of interest.  For a variable $s \in 
V$, let $N(s)$ denote the neighbors of $s$.  Then we say that a 
distribution $p$ is \emph{Markov with respect to $G$} if it meets the 
local Markov property: for any two variables $s,t \in V$, 
the variable $s$ is independent of $t$ 
conditioned on its neighbors $N(s)$. 
Intuitively, this means that the neighbors of $s$ contain all of the information 
necessary to predict its value. 
 
\begin{figure}
\resizebox{1.5in}{!}{\includegraphics{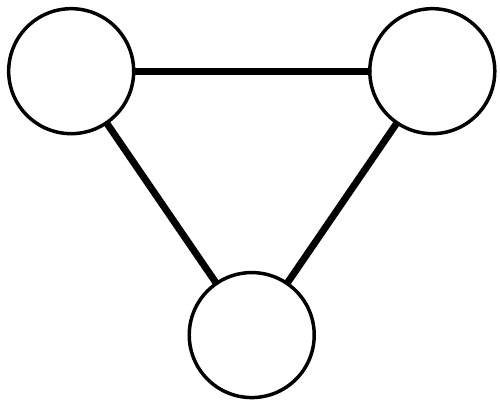}} 
\hfil 
\resizebox{1.5in}{!}{\includegraphics{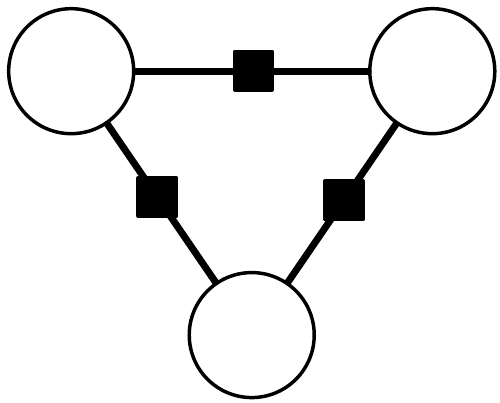}} 
\hfil 
\resizebox{1.5in}{!}{\includegraphics{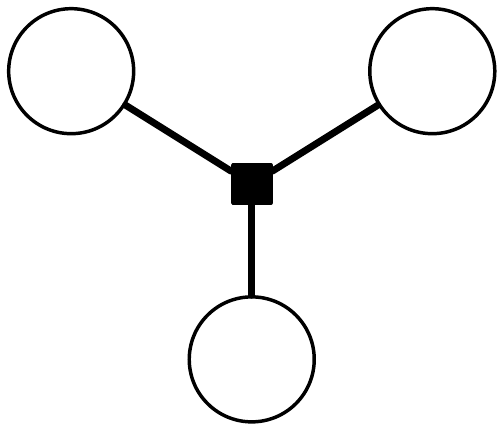}} 
\caption{A Markov network with an ambiguous factorization.  Both of the factor graphs at right factorize according to the Markov network at left.}\label{fig:bk:3var} 
\end{figure} 
Given a factorization of a distribution $p$ as in 
\eq{sm:eqn:factorization}, an equivalent Markov network can be 
constructed by connecting all pairs of variables that share a local 
function.  It is straightforward to show that $p$ is Markov with 
respect to this graph, because the conditional distribution $p(x_s | 
\bx_{N(s)})$ that follows from~(\ref{sm:eqn:factorization}) is a 
function only of variables that appear in the Markov blanket.   
In other words, if $p$ factorizes according to $G$, then $p$ is Markov 
with respect to $G$. 

The converse direction also holds, as long as $p$ is strictly positive. 
This is stated in the following classical result \citep{hammersley:clifford,besag74spatial}: 
 
\begin{theorem}[Hammersley-Clifford] 
  Suppose $p$ is a strictly positive distribution, and $G$ is an 
  undirected graph that indexes the domain of $p$.  Then $p$ is Markov 
  with respect to $G$ if and only if $p$ factorizes according to $G$. 
\end{theorem} 
 
A Markov network has an undesirable ambiguity from the factorization 
perspective, however.  Consider the three-node Markov network in 
\fig{fig:bk:3var} (left).  Any distribution that factorizes as 
$p(x_1,x_2,x_3) \propto f(x_1, x_2, x_3)$ for some positive function $f$ 
is Markov with respect to this graph. 
However, we may wish to use a more restricted parameterization, 
where $p(x_1, x_2, x_3) \propto f(x_1,x_2) g(x_2,x_3) h(x_1, x_3)$. 
This second model family is smaller,
and therefore may be more amenable to parameter estimation. 
But the Markov network formalism cannot distinguish between 
these two parameterizations. 
In order to state models more precisely, the factorization 
\eq{sm:eqn:factorization} can be represented directly 
by means of a  
\emph{factor graph} \citep{kschischang01factor}.  A factor graph is a 
bipartite graph $G = (V, F, E)$ in which a variable node $v_s \in V$ 
is connected to a factor node $\Psi_a \in F$ if $v_s$ is an argument 
to $\Psi_a$.  An example of a factor graph is shown graphically in 
\fig{fig:naive-bayes-gm} (right).  In that figure, the circles are 
variable nodes, and the shaded boxes are factor nodes.  Notice that, unlike the undirected
graph, the factor graph depicts the factorization of the model unambiguously.
 
\subsection{Directed Models} 
 
Whereas the local functions in an undirected model need not have a 
direct probabilistic interpretation, 
a \emph{directed graphical model} 
describes how a distribution factorizes into local conditional probability  
distributions. 
Let $G = (V, E)$ be a directed acyclic graph, in which  $\pi(v)$ are the parents of $v$ in $G$. A directed graphical model is a family of distributions that factorize as: 
\begin{equation} 
p(\ys,\xs) = \prod_{v \in V} p(y_v | \by_{\pi(v)}). 
\end{equation} 
It can be shown by structural induction on $G$ that $p$ is properly 
normalized.  Directed models can be thought of as a kind of factor
graph, in which the individual factors are locally normalized in a
special fashion so that globally $Z=1$.  Directed models are often used as generative models, 
as we explain in Section~\ref{bk:sec:generative-discriminative}. 
An example  
of a directed model is the naive Bayes model~\eqref{eqn:naive-bayes}, which is depicted graphically in \fig{fig:naive-bayes-gm} (left). 

\section{Generative versus Discriminative Models}
\label{sec:gendisc}
 
In this section we discuss several examples applications of simple graphical models to 
natural language processing.  Although these examples are well-known, 
they serve both to clarify the definitions in the previous section, and to illustrate some ideas that will arise again 
in our discussion of conditional random fields. 
We devote special attention to the hidden Markov model (HMM),  
because it is closely related to the linear-chain CRF. 
 
\subsection{Classification} 
\label{sec:classification} 
 
\begin{figure}[tbp] 
\centering    
\hspace{-0.5in} 
   \resizebox{1.5in}{!}{\includegraphics{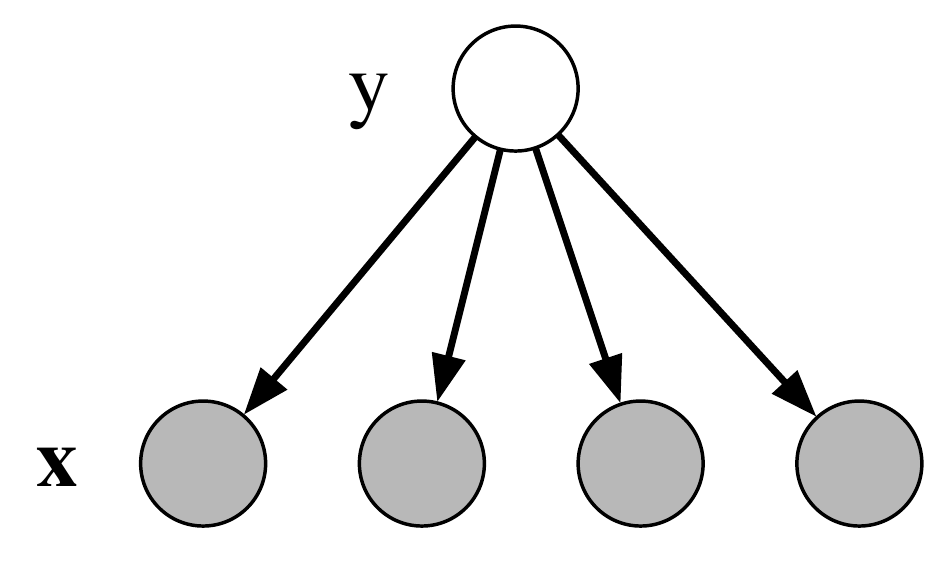}} 
   \hspace{0.5in} 
   \resizebox{1.5in}{!}{\includegraphics{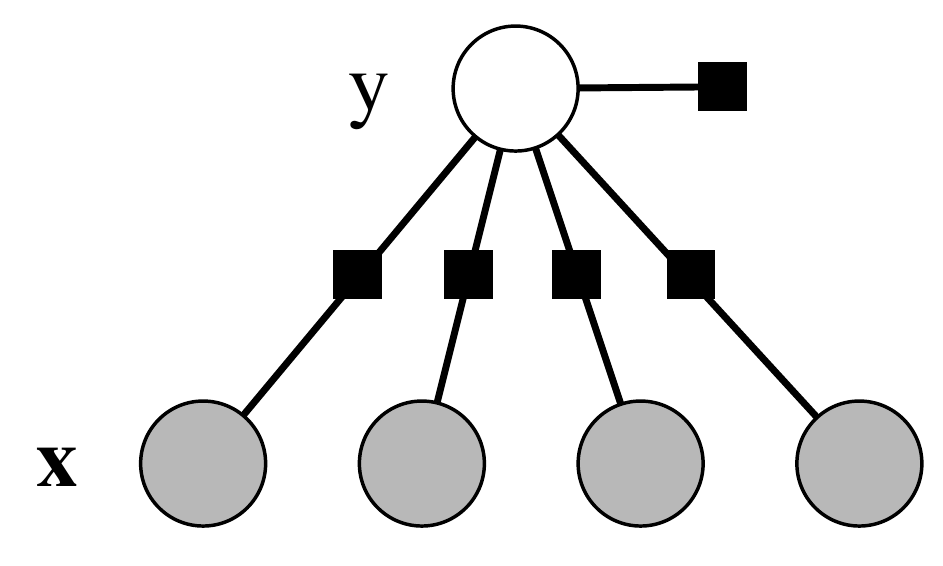}} 
    \caption{The naive Bayes classifier, as a directed model (left), 
    and as a factor graph (right).} 
   \label{fig:naive-bayes-gm} 
   \label{fig:naive-bayes-fg} 
\end{figure} 
 
First we discuss the problem of \emph{classification}, that 
is, predicting a single discrete class variable $y$ given a vector 
of features $\xs = (x_1, x_2, \ldots, x_K)$. 
One simple way to accomplish this is 
to assume 
that once the class label is known, all  
the features are independent.  The resulting classifier 
is called the \emph{naive Bayes classifier}. 
It is based on a joint probability model of the form: 
\begin{equation} 
p(y,\xs) = p(y) \prod_{k=1}^K p(x_k | y).\label{eqn:naive-bayes} 
\end{equation} 
This model can be described by the directed model shown 
in \fig{fig:naive-bayes-gm} (left).  We can also write this 
model as a factor graph, by defining a factor $\Psi(y) = p(y)$, and a factor $\Psi_k (y, x_k) = p(x_k | y)$ 
for each feature $x_k$. 
This factor graph is shown in \fig{fig:naive-bayes-fg} (right). 
 
Another well-known classifier that is naturally represented 
as a graphical model is  
logistic regression (sometimes known as the \emph{maximum 
entropy classifier} in the NLP community).  In statistics, 
this classifier is motivated by the assumption that the log 
probability, $\log p(y|\xs)$, of each class is a linear function of $\xs$, plus 
a normalization constant.  This leads to the conditional 
distribution: 
\begin{equation} 
p(y | \xs) = \frac{1}{Z(\xs)} \exp\left\{\theta_y + \sum_{j=1}^K \theta_{y,j} x_j\right\},\label{eqn:lr-1perclass} 
\end{equation} 
where $Z(\xs) = \sum_y \exp\{\theta_y + \sum_{j=1}^K \theta_{y,j} x_j\}$ is  
a normalizing constant, and $\theta_y$ is a bias weight that acts 
like $\log p(y)$ in naive Bayes.  Rather than using one weight vector 
per class, as in \eq{eqn:lr-1perclass}, we can use a different 
notation in which a single set of weights is shared across all the classes.  The trick is to define a set of \emph{feature functions} that are nonzero only for a single class.  To do this, 
the feature functions can be defined as 
$f_{y',j} (y, \xs) = \Ind{y' = y} x_j$ for the feature weights and 
$f_{y'} (y, \xs) = \Ind{y' = y}$ for the bias weights.  Now we 
can use $f_k$ to index each feature function $f_{y',j}$, 
and $\theta_k$ to index its corresponding weight $\theta_{y',j}$. 
Using this notational trick, the logistic 
regression model becomes: 
\begin{equation} 
p(y | \xs) = \frac{1}{Z(\xs)} \exp\left\{\sum_{k=1}^K \theta_k 
f_k(y, \xs) \right\}.\label{eqn:lr} 
\end{equation} 
We introduce this notation because it  
mirrors the notation for conditional random fields that we will
present later. 
 
 
\subsection{Sequence Models} 
\label{sec:bk:sequence} 
 
Classifiers predict only a single class variable, 
but the true power of graphical models lies in their 
ability to model 
many variables that are interdependent. 
In this section, we discuss perhaps the simplest form of 
dependency, in which the output variables are arranged in a sequence. 
To motivate this kind of model, we discuss  
an application from natural language processing, the task of 
\emph{named-entity recognition} (NER). 
NER is the problem of identifying and classifying proper names in text, including locations, such as \emph{China}; people, such as  
\emph{George Bush}; and organizations, such as the \emph{United 
Nations}. 
The named-entity recognition task is,  
given a sentence, to segment which words are part of 
entities, and to classify each entity by type 
(person, organization, location, and so on). 
The challenge of this problem is that 
many named entities are too rare to appear even in a large 
training set,  and therefore the system must identify  
them 
based only on context.   
 
One approach to NER is 
to classify each word independently as one of either 
\etype{Person}, \etype{Location}, \etype{Organization}, 
or \etype{Other} (meaning not an entity).  
The problem with this approach is that it assumes 
 that given the input,  
all of the named-entity labels are independent. 
In fact, the named-entity labels of neighboring words 
are dependent; for example, while \emph{New York} is a location,  
\emph{New York Times} is an organization.  One way to relax this independence assumption 
is to arrange the output variables in a linear chain.  This is the 
approach taken by the hidden Markov model (HMM) \citep{rabiner89hmm}.  An HMM  
models a sequence of observations $X = \{ x_t \}_{t=1}^\sT$  
by assuming that there is an underlying sequence of \emph{states} 
$Y = \{ y_t \}_{t=1}^\sT$ drawn from a finite state set $S$. 
In the named-entity example, each observation $x_t$ is 
 the identity of the word at position $t$, and each state 
$y_t$ is the named-entity label, that is, one of  
the entity types 
\etype{Person}, \etype{Location}, \etype{Organization}, 
and \etype{Other}. 
  
To model the joint distribution $p(\ys,\xs)$ tractably, an 
HMM makes two independence assumptions. 
First, it assumes that each state depends only on its 
immediate predecessor, that is, each state $y_t$ is independent of 
all its ancestors $y_1, y_2, \ldots, y_{t-2}$ given 
the preceding state $y_{t-1}$.  Second, it also assumes 
that each observation variable $x_t$ depends only on the 
current state $y_t$. 
With these assumptions, we can specify an HMM 
using three probability distributions: first, 
the distribution $p(y_1)$ over initial states; 
second, the transition distribution $p(y_t | y_{t-1})$; 
and finally, the observation distribution $p(x_t | y_t)$. 
That is, the joint probability 
of a state sequence $\ys$ and an observation sequence $\xs$ 
factorizes as 
\begin{equation} 
p(\ys, \xs) = \prod_{t=1}^\sT p(y_t | y_{t-1}) p(x_t | y_t)\label{sm:eqn:hmm}, 
\end{equation} 
where, to simplify 
notation, we write the initial state distribution 
$p(y_1)$ as $p(y_1 | y_0)$.   
In natural language processing, HMMs have been used for 
sequence labeling tasks such as part-of-speech tagging, 
named-entity recognition, and information extraction. 
 
 

\subsection{Comparison}
\label{bk:sec:generative-discriminative} 
 
Of the models described in this section, two are generative (the naive
Bayes and hidden Markov models)
and one is discriminative (the logistic regression model).
In a general, \emph{generative models}
are models of the joint distribution $p(y,\xs)$, and like naive Bayes have the form
$p(\by) p(\bx | \by)$.  In other words, they describe how the output is
probabilistically generated as a function of the input.
\emph{Discriminative models}, on the other hand, focus solely 
on the conditional distribution $p(y | \xs)$.  In this section, 
we discuss the differences between generative and discriminative 
modeling, and the potential advantages of discriminative modeling. 
For concreteness, we focus on the 
examples of naive Bayes and logistic regression, 
but the discussion in this section 
applies equally as well to the differences between arbitrarily structured generative 
models and conditional random fields. 
 
 
The main difference is that a conditional distribution $p(\ys|\xs)$ 
does not include a model of $p(\xs)$, which is not needed 
for classification anyway. 
The difficulty in modeling $p(\xs)$ is that it often contains 
many 
highly dependent features that are difficult  
to model.  
For example, in named-entity recognition, 
an HMM relies on only one feature, 
the word's identity. 
But many words, especially proper names, 
will not have occurred in the training set, 
so the word-identity feature is uninformative. 
To label unseen words, 
we would like to exploit other features of a word, 
such as its capitalization, 
its neighboring words, its prefixes and suffixes,  
its membership in predetermined 
lists of people and locations, and so on.

 The principal advantage of discriminative modeling is 
 that it is better suited to including rich, overlapping features. 
 To understand this, 
 consider the 
 family of naive Bayes distributions \eq{eqn:naive-bayes}. 
 This is a family of joint distributions whose conditionals 
 all take the ``logistic regression form'' \eq{eqn:lr}. 
But there are many other joint models,  
some with complex dependencies among $\xs$, whose conditional distributions also have the form \eq{eqn:lr}. 
By modeling the conditional distribution directly, 
we can remain agnostic about the form of $p(\xs)$. 
CRFs make independence 
assumptions among $\ys$, and assumptions about how the $\ys$ can depend on $\xs$, but not among $\xs$. 
This point can also be understood graphically: Suppose that we have a factor graph representation for the joint distribution $p(\ys, \bx)$.  
If we then construct a graph for the conditional distribution $p(\ys | \bx)$, any factors that depend only on
$\bx$ vanish from the graphical structure for the conditional distribution.  They are irrelevant to the conditional because they are constant with respect to $\ys$.

To include interdependent features in a generative 
model, we have two choices: enhance the model 
to represent dependencies among the inputs, or make simplifying 
independence assumptions, such as the naive Bayes assumption. 
The first approach, enhancing the model, is often difficult to 
do while retaining tractability.  For example, it is hard to imagine how to model 
the dependence between the capitalization of 
a word and its suffixes, nor do we particularly wish to do so,  
since we always observe the test sentences anyway. 
The second approach---to include a large number of dependent features in a generative model, 
but to include  independence assumptions among them---is possible, and in some domains can work well. 
But it can also be problematic because the independence assumptions can hurt 
performance.  For example, although the naive Bayes 
classifier performs well in document 
classification, it performs worse on average 
across a range of applications 
than logistic regression 
\citep{caruana05empirical}. 
 
Furthermore, naive Bayes  can produce poor probability estimates.
As an illustrative example, imagine training naive Bayes on a data set 
in which all the features are repeated, that is, 
$\xs = (x_1, x_1, x_2, x_2, \ldots, x_K, x_K)$.  This will increase the 
confidence of the naive Bayes probability estimates, even though no new 
information has been added to the data.  Assumptions like 
naive Bayes can be especially 
problematic when we generalize to sequence models, because 
inference essentially combines evidence from different parts 
of the model. 
If probability estimates of the label at each sequence position 
are overconfident, it might be difficult to combine 
them sensibly. 
 
The difference between 
naive Bayes and logistic regression is due \emph{only} 
to the fact 
that the first is generative and the second discriminative; 
the two classifiers are, for discrete input, 
identical in all other respects. 
Naive Bayes and logistic regression consider the same hypothesis space, 
in the sense that any logistic regression 
classifier can be converted into a naive Bayes classifier with the 
same decision boundary, and vice versa. 
Another way of saying this is that  
the naive Bayes model \eq{eqn:naive-bayes} 
defines the same family of distributions as 
the logistic regression model \eq{eqn:lr}, 
if we interpret it generatively as 
\begin{equation} 
p(y,\xs) = \frac{\exp\left\{\sum_k \theta_k f_k (y, \xs)\right\}}{\sum_{\tilde{y},\tilde{\xs}} \exp\left\{\sum_k \theta_k f_k (\tilde{y}, \tilde{\xs})\right\}}.\label{eqn:lr-generative} 
\end{equation} 
This means that if the naive Bayes model \eq{eqn:naive-bayes} 
is trained to maximize the conditional likelihood, 
we recover the same classifier as from logistic regression. 
Conversely, if the logistic regression model is interpreted 
generatively, as in \eq{eqn:lr-generative}, and is trained 
 to maximize the joint 
likelihood $p(y,\xs)$, 
then we recover the same classifier as from naive Bayes. 
In the terminology of \citeauthornum{ng02discriminative}, naive 
Bayes and logistic regression form a 
\emph{generative-discriminative pair}.  
For a recent theoretical perspective on generative and discriminative models, see \citeauthornum{liang08asymptotic}.

\begin{figure}[tbp] 
   \centering 
   \resizebox{!}{2.1in}{\includegraphics{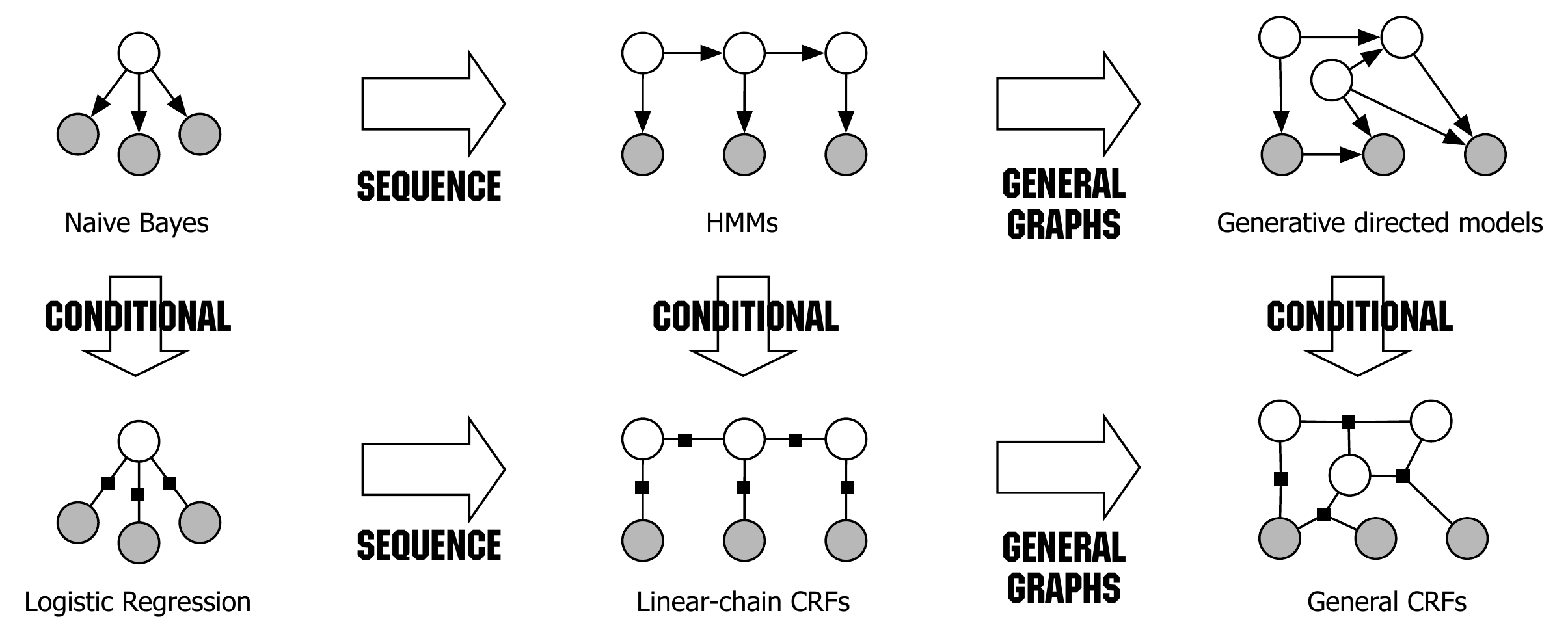}} 
   \vspace{0.05in} 
   \caption{Diagram of the relationship between naive Bayes, logistic regression, HMMs, linear-chain CRFs, generative models, and general 
   CRFs.   
   } 
   \label{fig:analogy} 
\end{figure} 
 
One perspective for gaining insight into the difference 
between generative and discriminative modeling is 
 due to \citeauthornum{minka05discriminative}.  Suppose we  
 have a generative model $p_g$ with parameters $\theta$.  By 
 definition, this takes the form 
\begin{equation} 
p_g(\ys,\xs; \theta) = p_g(\ys; \theta) p_g(\xs | \ys; \theta). 
\end{equation} 
But we could also rewrite $p_g$ using Bayes rule as 
\begin{equation} 
p_g(\ys,\xs; \theta) = p_g (\xs; \theta) p_g (\ys | \xs; \theta),\label{sm:minka:generative} 
\end{equation} 
where $p_g (\xs; \theta)$ and $p_g (\ys | \xs; \theta)$ are computed by inference, i.e., $p_g (\xs; \theta) = \sum_\ys p_g (\ys,\xs; \theta)$ and $p_g (\ys | \xs; \theta) =  p_g (\ys,\xs; \theta) / p_g (\xs; \theta)$.    
 
Now, compare this generative model to a discriminative 
model over the same family of joint distributions. 
To do this, we define a prior $p(\xs)$ over inputs, 
such that $p(\xs)$ could have arisen from $p_g$ 
with some parameter setting.  That is, $p(\xs) = p_c(\xs;\theta') =\sum_{\ys} p_g (\ys,\xs | \theta')$.  We combine this with a conditional 
distribution $p_c (\ys | \xs; \theta)$ that could also 
have arisen from $p_g$, that is, $p_c (\ys | \xs; \theta) 
= p_g (\ys, \xs; \theta) / p_g (\xs; \theta)$. 
Then the resulting distribution is 
\begin{equation} 
p_c (\ys,\xs) = p_c(\xs; \theta') p_c (\ys | \xs; \theta). 
\label{sm:minka:discriminative} 
\end{equation} 
By comparing \eq{sm:minka:generative} with \eq{sm:minka:discriminative}, 
it can be seen that the conditional approach 
has more freedom to fit the data, 
because it does not require that $\theta = \theta'$. 
Intuitively, because  
the parameters $\theta$  
in \eq{sm:minka:generative} 
are used in both the input distribution and the conditional, a good set of parameters must represent both well,  
potentially at the cost of trading off accuracy on $p(\ys|\xs)$, the distribution we care about, for accuracy on $p(\xs)$, which we care less about.  On the other hand, this added freedom brings about an increased risk of overfitting the training data, and generalizing worse on unseen data. 
 
To be fair, however, generative models have several advantages of their own. 
First, generative models can be more natural for handling latent variables, partially-labeled 
data, and unlabelled data.  In the most extreme case, 
when the data is entirely unlabeled, generative models can be applied in an 
unsupervised fashion, 
whereas unsupervised learning in discriminative models is less natural
and is still an active area of research.

Second, on some data a generative model can perform better 
than a discriminative model, intuitively because the input model $p(\bx)$ 
may have a smoothing effect on the conditional.  \citeauthornum{ng02discriminative} 
argue that this effect is especially pronounced when the data set is small. 
For any particular data set, it is impossible to predict in advance whether 
a generative or a discriminative model will perform better. 
Finally, sometimes either the problem suggests a natural generative model, 
or the application requires the ability to predict both future inputs and future outputs, making a generative model 
preferable. 

Because a generative model takes the form $p(\by,\bx) = p(\by) p(\bx | \by)$, 
it is often natural to represent a generative model by a directed graph
in which in outputs $\y$ topologically precede the inputs. 
Similarly, we will see that it is often natural to represent a discriminative model by a undirected graph,
although this need not always be the case.

The relationship between naive Bayes 
and logistic regression 
mirrors the relationship between HMMs and 
linear-chain CRFs. 
Just as naive Bayes and logistic regression are a 
generative-discriminative pair, there is a 
discriminative analogue to the hidden Markov model, 
and this analogue is a particular special case of 
conditional random field, as we explain in the next section. 
This analogy between naive Bayes, logistic regression, 
generative models, and conditional random fields 
is depicted in \fig{fig:analogy}. 

\section{Linear-chain CRFs}
\label{sec:lc-crf-defn}

\begin{figure}[tbp] 
   \centering 
   \hspace{-0.5in} 
   \resizebox{2.5in}{!}{\includegraphics{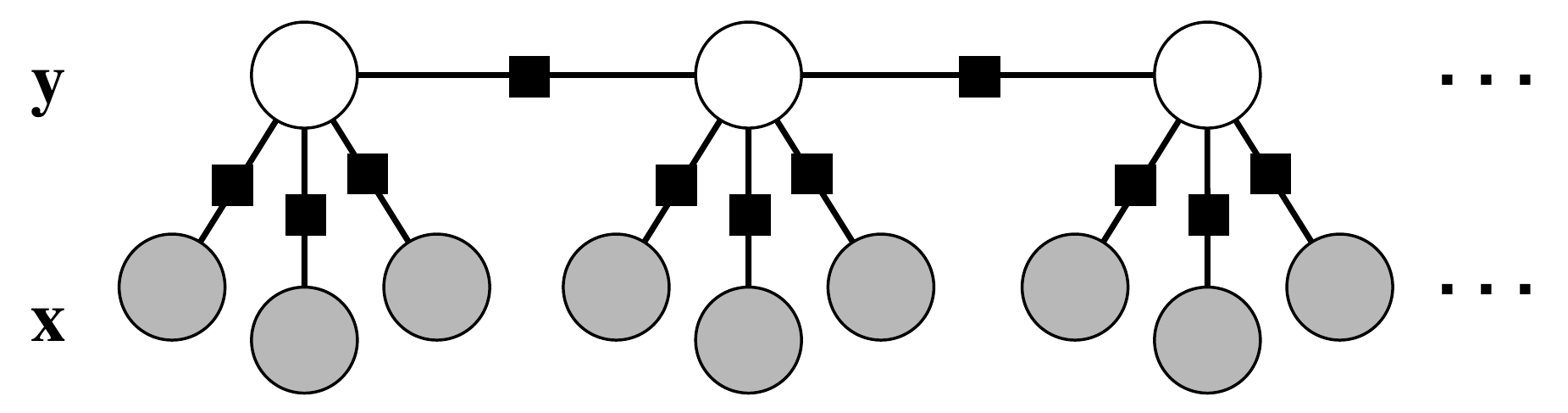}} 
   \vspace{0.1in} 
   \caption{Graphical model of an HMM-like linear-chain CRF.} 
   \label{fig:hmmlike-linear-crf} 
\end{figure} 
 
\begin{figure}[tbp] 
   \centering 
   \hspace{-0.5in} 
   \resizebox{2.5in}{!}{\includegraphics{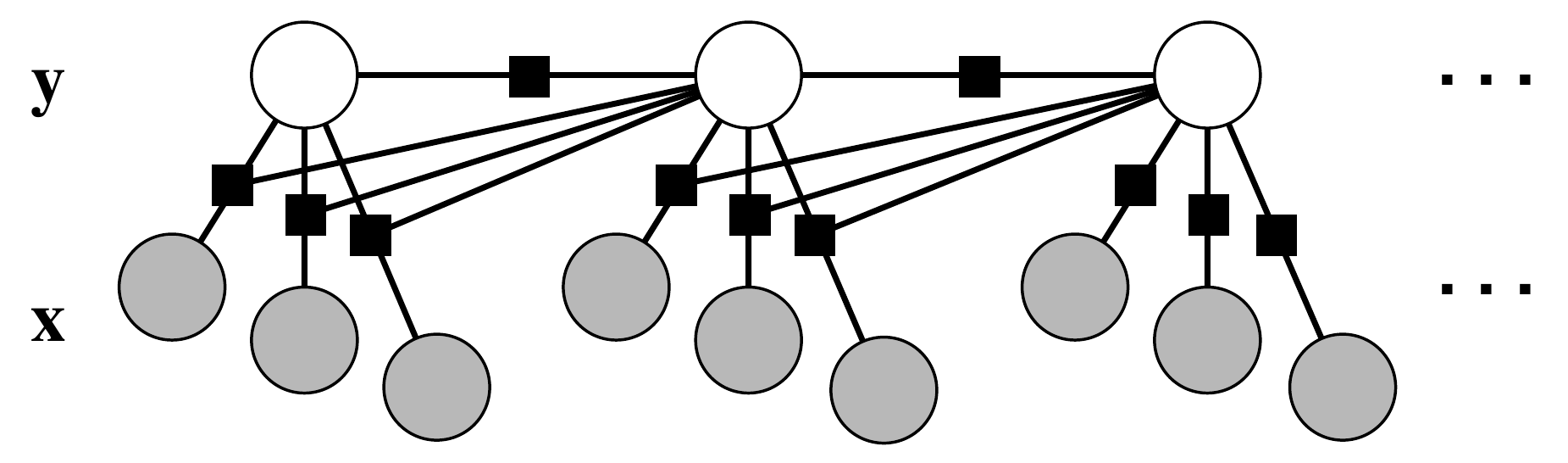}} 
   \vspace{0.1in} 
   \caption{Graphical model of a linear-chain CRF 
   in which the transition score depends on the  
   current observation.} 
   \label{fig:linear-crf} 
\end{figure} 
 
To motivate our introduction of  
linear-chain conditional random fields, we begin by considering  
the conditional distribution $p(\ys | \xs)$ that  
follows from the joint distribution $p(\ys, \xs)$ 
of an HMM.  The key point is that this conditional 
 distribution is in fact a conditional random field with a particular choice of feature functions.  
 
First, we rewrite the HMM joint \eq{sm:eqn:hmm} 
in a form that is more amenable to generalization.  This is 
\begin{equation} 
p(\ys, \xs) = \frac{1}{Z} \prod_{t=1}^T \exp\left\{ \sum_{i,j \in S} \theta_{ij} \Ind{y_t = i} \Ind{y_{t-1} = j} + \sum_{i \in S} \sum_{o \in O} \mu_{oi} \Ind{y_t = i} \Ind{x_t = o} \right\},\label{sm:eqn:hmm-as-exp} 
\end{equation}
where $\theta = \{ \theta_{ij}, \mu_{oi} \}$ are the real-valued parameters of
the distribution and $Z$ is a
normalization constant chosen so the distribution sums to
one.\footnote{Not all choices of $\theta$ are valid, because the
  summation defining $Z$, that is, $Z = \sum_{\ys} \sum_{\xs}
  \prod_{t=1}^T \exp\left\{ \sum_{i,j \in S}
    \theta_{ij} \Ind{y_t = i} \Ind{y_{t-1} = j} + 
    \sum_{i \in S} \sum_{o \in O} \mu_{oi} \Ind{y_t = i} \Ind{x_t = o}
  \right\}$, might not converge.  An example of this is a model with one state where
  $\theta_{00} > 0$.  
This issue is typically not an issue for CRFs, because in a CRF the summation within $Z$
is usually over a finite set.} 
It can be seen that  \eq{sm:eqn:hmm-as-exp} describes exactly the class of HMMs.
Every HMM can be written in this form by setting $\theta_{ij} = \log
p(y' = i | y = j)$ and $\mu_{oi} = \log p(x = o | y = i)$.
The converse direction is more complicated, and not relevant for our
purposes here.  The main point is that despite this added flexibility in the parameterization
 \eq{sm:eqn:hmm-as-exp}, we have not added any distributions to the family. 

We can write \eq{sm:eqn:hmm-as-exp} more compactly by introducing  
the concept of \emph{feature functions}, just as we did for 
logistic regression in \eq{eqn:lr}.  Each  
feature function has the form 
$f_k (y_t, y_{t-1}, x_t)$.   
 In order to duplicate \eq{sm:eqn:hmm-as-exp}, there needs to be one feature $f_{ij} (y,y',x) = \Ind{y = i} \Ind{y' = j}$ for each transition 
 $(i,j)$ and one feature $f_{io} (y,y',x) = \Ind{y = i}\Ind{x=o}$ for each state-observation pair $(i, o)$. 
We refer to a feature function generically as $f_k$, where $f_k$ ranges over both all of the $f_{ij}$ and all of the $f_{io}$.
Then we can write an HMM as: 
\begin{equation} 
p(\ys, \xs) = \frac{1}{Z} \prod_{t=1}^T \exp\left\{\sum_{k=1}^K \theta_k f_k (y_t, y_{t-1}, x_t) \right\}.\label{sm:eqn:hmm-as-exp-compact} 
\end{equation} 
Again, 
equation \eq{sm:eqn:hmm-as-exp-compact} defines exactly 
the same family of distributions as \eq{sm:eqn:hmm-as-exp}, 
and therefore as the original HMM equation \eq{sm:eqn:hmm}. 
  
The last step is to write the conditional distribution $p(\ys|\xs)$  
that results from the HMM \eq{sm:eqn:hmm-as-exp-compact}.  This is 
\begin{equation} 
p(\ys| \xs) = \frac{p(\ys, \xs)}{\sum_{\ys'} p(\ys', \xs)} =
\frac{\prod_{t=1}^T \exp\left\{\sum_{k=1}^K \theta_k f_k (y_t,
    y_{t-1}, x_t) \right\}}{\sum_{{\ys'}} \prod_{t=1}^T \exp\left\{\sum_{k=1}^K \theta_k f_k (y'_t,y'_{t-1}, x_t) \right\}}. 
\label{eqn:hmm-as-crf} 
\end{equation} 
This conditional distribution \eq{eqn:hmm-as-crf} is a particular kind 
of linear-chain CRF, namely, one that includes features only 
for the current word's identity. 
But many other linear-chain CRFs use richer 
features of the input, such as prefixes and suffixes  
of the current word, the identity of surrounding words, and so on.
Fortunately, this extension requires little change to our existing
notation.  We simply allow the feature functions
to be more general than indicator functions of the word's identity.  This leads to the general 
definition of linear-chain CRFs:
 
\begin{definition} 
Let $Y, X$ be random vectors, $\theta = \{ \theta_k \} \in \Re^K$ be  a parameter vector, and $\{ f_k (y, y', \xs_t) \}_{k=1}^K$ be a 
set of real-valued feature functions.  Then a  
\emph{linear-chain conditional random field} 
is a distribution $p(\ys|\xs)$ that takes the form 
\begin{equation} 
p(\ys|\xs) = \frac{1}{Z(\xs)} \prod_{t=1}^T \exp\left\{\sum_{k=1}^K \theta_k f_k (y_t, y_{t-1}, \xs_t) \right\}\label{sm:eqn:lc-crf}, 
\end{equation} 
where $Z(\xs)$ is an instance-specific normalization function 
\begin{equation} 
Z(\xs) = \sum_{\ys} \prod_{t=1}^T \exp\left\{\sum_{k=1}^K \theta_k f_k (y_t, y_{t-1}, \xs_t) \right\}. 
\end{equation} 
\label{def:lc-crf} 
\end{definition} 
 
We have just seen that if the joint $p(\ys,\xs)$ factorizes as an HMM,  
then the associated conditional distribution $p(\ys|\xs)$ is 
a linear-chain CRF.  This HMM-like CRF is pictured in \fig{fig:hmmlike-linear-crf}.  Other types of linear-chain CRFs 
are also useful, however.  For example, typically in an HMM, a 
transition from state $i$ to state $j$ receives the 
same score, $\log p(y_t = j | y_{t-1} = i)$, regardless 
of the input.  In a CRF, we can allow 
the score of the transition $(i,j)$ 
to depend on the current observation vector, 
simply by adding  
a feature $\Ind{y_t = j}\Ind{y_{t-1} = 1}\Ind{x_t = o}.$ 
 A CRF 
with this kind of transition feature, 
which is commonly used in text applications, 
is pictured in \fig{fig:linear-crf}. 
 
To indicate in the definition 
of linear-chain CRF that each feature function can depend 
on observations from any time step, we have  
written  the observation argument to $f_k$ as a vector $\xs_t$,  
which should 
be understood as containing all the components of the global observations 
$\xs$ that are needed for computing features at time $t$. 
For example, if the CRF uses the next word $x_{t+1}$ 
as a feature, then the feature vector $\xs_t$ is assumed to  
include the identity of word $x_{t+1}$. 
 
Finally, note that the normalization 
constant $Z(\xs)$ sums over 
all possible state sequences, 
an exponentially large number of terms.  Nevertheless, it can be computed efficiently by forward-backward, as we explain in \sec{sm:sec:lc-inference}.  
 
\section{General CRFs} 
\label{sec:mdl:crf-general}

Now we present the general definition of a conditional 
random field, as it was originally introduced \cite{lafferty01crf}.
The generalization from linear-chain CRFs to general 
CRFs is fairly straightforward.  We simply move from 
using a linear-chain factor graph to a more general 
factor graph, and from forward-backward to more 
general (perhaps approximate) inference algorithms.

\begin{definition} 
Let $G$ be a factor graph over $Y$.  Then $p(\ys|\xs)$ is a conditional random field if for any fixed $\xs$, 
the distribution $p(\ys|\xs)$ 
factorizes according to $G$. 
\end{definition} 
 
Thus, every conditional distribution $p(\ys|\xs)$ is a CRF for some, perhaps trivial, factor graph. 
If $F = \{ \Psi_a \}$ is the set of factors in $G$, 
and each factor takes the exponential family form 
\eq{sm:eqn:exponential-factor}, 
 then the conditional distribution can be written as 
\begin{equation} 
p(\ys | \xs) = \frac{1}{Z(\xs)} \prod_{\Psi_{\! A} \in G} \exp \left\{\sum_{k=1}^{K(A)} \theta_{ak} f_{ak} (\ys_a, \xs_a) \right\}. 
\end{equation} 
In addition, 
practical models rely extensively on parameter tying.  For example, in the linear-chain case, often the 
same weights are used for the factors $\Psi_t (y_t, y_{t-1}, \xs_t)$ at each time step. 
To denote this, we partition 
the factors of $G$ into $\smCalC = \{ C_1, C_2, \ldots C_P \}$, where
each $C_p$ is a \emph{clique template} whose parameters are tied.
This notion of clique template generalizes that in
\citeauthornum{taskar02rmn}, \citeauthornum{sutton04dcrf},
\citeauthornum{richardson05markov}, and \citeauthornum{mccallum09factorie}. 
  Each clique template $C_p$ is a set of factors which has a corresponding set of sufficient statistics $\{ f_{pk} (\xs_p, \ys_p) \}$ and parameters $\theta_p \in \Re^{K(p)}$. Then the CRF can be written as  
\begin{equation} 
p(\ys|\xs) = \frac{1}{Z(\xs)} \prod_{C_p \in \smCalC} \prod_{\Psi_c \in C_p} \Psi_c (\xs_c, \ys_c; \theta_p), 
\end{equation} 
where each factor is parameterized as 
\begin{equation} 
\Psi_c (\xs_c, \ys_c; \theta_p) = \exp\left\{\sum_{k=1}^{K(p)} \theta_{pk} f_{pk} (\xs_c, \ys_c) \right\}, \label{eq:crf-factors}
\end{equation} 
and the normalization function is 
\begin{equation} 
Z(\xs) = \sum_{\ys} \prod_{C_p \in \smCalC} \prod_{\Psi_c \in C_p} \Psi_c(\xs_c, \ys_c; \theta_p). 
\end{equation} 
This notion of clique template specifies both repeated structure and parameter tying in the model.
For example, in a linear-chain conditional random field, typically one clique template  
$C_0 = \{ \Psi_t(y_{t}, y_{t-1}, \xs_t) \}_{t=1}^{\sT}$ is used for the entire 
network, so $\smCalC = \{ C_0 \}$ is a singleton set.  If instead we want each factor $\Psi_t$ to have a separate set of parameters,
this would be accomplished using $T$ templates, by taking $\smCalC = \{ C_t \}_{t=1}^{\sT}$, where
$C_t = \{ \Psi_t(y_{t}, y_{t-1}, \xs_t) \}$.
Both the set of clique templates and the number of outputs can depend on the input $\bx$; for example, 
to model images, we may use different clique templates at different scales depending on the results of an algorithm for finding points of interest.
 
One of the most important considerations in defining a general CRF lies in specifying the repeated structure and parameter tying.
A number of formalisms have been proposed to specify the clique templates.  
For example, \emph{dynamic conditional random fields} \citep{sutton04dcrf} 
are sequence models which allow multiple labels at each 
time step, rather than single label, in a manner analogous to dynamic Bayesian networks.
Second, \emph{relational Markov networks} \citep{taskar02rmn} are a type of general CRF in which the graphical structure and parameter tying are determined by an SQL-like syntax.   
\emph{Markov logic networks} \citep{richardson05markov,singla05discriminative}  
use logical formulae to specify the scopes of local functions in an undirected model.  Essentially, there is a set of parameters for 
each first-order rule in a knowledge base.  The logic portion of an MLN can be viewed as essentially a programming convention for
specifying the repeated structure and parameter tying of an undirected model.
\emph{Imperatively defined factor graphs} \citep{mccallum09factorie} use the full expressivity of Turing-complete functions
to define the clique templates, specifying both the structure of the model and the sufficient statistics $f_{pk}$.
These functions have the flexibility to employ advanced programming ideas including recursion, arbitrary search, lazy evaluation, and memoization.


\section{Applications of CRFs}
\label{chp:applications}

CRFs have been applied to a variety of domains, 
including text processing, computer vision, 
and bioinformatics. 
One of the first large-scale applications of 
CRFs was by \citeauthornum{sha03shallow}, who matched state-of-the-art 
performance on segmenting noun phrases in text. 
Since then, linear-chain CRFs have been 
applied to many problems in natural language processing, 
including named-entity recognition 
\citep{li03conll}, feature induction for NER 
\citep{mccallum03ifcrf}, 
shallow parsing \citep{sha03shallow,sutton07dcrf},
identifying protein names in biology abstracts \citep{settles05abner}, 
segmenting addresses in Web pages \citep{culotta04extracting}, 
information integration \citep{wick08ontology},
finding semantic roles in text \citep{RothYi05}, 
prediction of pitch accents \cite{gregory04crf},
phone classification in speech processing \citep{gunawardana05hidden},
identifying the sources of opinions \citep{choi05identifying}, 
word alignment in machine translation \citep{blunsom06acl},
citation extraction from research papers \citep{peng04ie},
extraction of information from tables in text documents \citep{pinto03table},
Chinese word segmentation \citep{peng04chinese},  
Japanese morphological analysis \citep{kudo04japanese}, 
and many others. 
 
In bioinformatics, CRFs have been applied 
to RNA structural alignment \citep{sato05rna} 
and protein structure prediction \citep{liu06segmentation}. 
Semi-Markov CRFs \citep{sarawagi05semi} add  
somewhat more flexibility in choosing features, by allowing features
functions to depend on larger segments of the input that depend on the output
labelling.
This can be useful for certain tasks in information extraction and 
especially bioinformatics. 
 
General CRFs have also been applied to several tasks in NLP. 
One promising application 
is to performing multiple labeling tasks simultaneously. 
For example, \citeauthornum{sutton04dcrf} show that a two-level dynamic 
CRF for part-of-speech tagging and noun-phrase 
chunking performs better than solving the tasks one at a time.  
Another application is to \emph{multi-label classification}, 
in which each instance can have multiple class labels. 
Rather than learning an independent classifier for 
each category, \citeauthornum{ghamrawi05collective} present a CRF that 
learns dependencies between the categories,  
resulting in improved classification performance. 
Finally, the skip-chain CRF \cite{sutton04skip}
is a general CRF that represents long-distance 
dependencies in information extraction. 
 
An interesting graphical CRF structure has 
been applied to the problem of proper-noun coreference, that 
is, of determining which mentions in a document, 
such as \emph{Mr. President} 
and \emph{he}, refer to the same underlying entity. 
 \citeauthornum{mccallum05wellner} learn a distance metric 
 between mentions using a fully-connected conditional 
 random field  
 in which inference corresponds to graph 
 partitioning. 
A similar model has been used 
to segment handwritten characters and diagrams 
 \citep{cowans05graphical,qi05diagram}.

In computer vision, several authors  
have used grid-shaped CRFs  
\citep{he04mcrfs,kumar03drf} for labeling and segmenting 
images.  Also, for recognizing objects, \citeauthornum{quattoni05conditional} 
use a tree-shaped CRF in which latent variables are designed to 
recognize characteristic parts of an object.

In some applications of CRFs, efficient dynamic programs 
exist even though the graphical model is difficult to specify. 
For example, \citeauthornum{mccallum05stringedit} learn the parameters 
of a string-edit model in order to discriminate between matching 
and nonmatching pairs of strings.  Also, there is 
work on using CRFs to learn distributions over the 
derivations of a grammar \citep{riezler02wsj,clark04parsing,viola05learning,finkel08efficient}. 

\section{Feature Engineering}
\label{sec:feature-engineering}
 
In this section we describe some ``tricks of the trade'' that involve
feature engineering.  Although these apply especially to language applications, they are also useful more generally. 
 
First, when the predicted variables are discrete, the features $f_{pk}$ of a clique template $C_p$ are ordinarily chosen to have a particular form: 
\begin{equation} 
f_{pk} (\ys_c, \xs_c) = \Ind{\ys_c = \tilde{\ys}_c} q_{pk}(\xs_c).\label{sm:eqn:feature-factorization} 
\end{equation} 
In other words, each feature is nonzero only for a single output configuration $\tilde{\ys}_c$, but as long as that constraint is met,  
then the feature value depends only on the input observation. 
Essentially, this means that we can think of our features as depending only on the input $\xs_c$, but that we have a separate set of  
weights for each output configuration. 
This feature representation is also computationally efficient,  
because computing each $q_{pk}$ may involve nontrivial text or image processing, and it need be evaluated only once for every feature that uses it. To avoid confusion, we refer to the functions $q_{pk}(\xs_c)$ as \emph{observation functions} rather than as features.  Examples of observation functions are ``word $x_t$ is capitalized'' and ``word $x_t$ ends in \emph{ing}''. 
 
This representation can lead to a large number of features, which can have significant memory and time requirements.  For example, to match state-of-the-art results on a standard natural language task,  \citeauthornum{sha03shallow} use 3.8 million features.  Many of these features 
always zero in the training data.  In particular, some observation functions $q_{pk}$ are nonzero for certain output configurations and zero for others.  This point can be confusing: One might think that such features can have no effect on the likelihood, but actually 
putting a negative weight on them causes an assignment that does not appear in the training data to become less likely, which improves the likelihood.
For this reason, including unsupported features typically results in better accuracy.  
In order to save memory, however, sometimes these \emph{unsupported features}, that is, those which never occur in the training data, are removed from the model.  

As a simple heuristic for getting some of the benefits of unsupported features with less memory, 
we have had success with an ad hoc technique for selecting a small set of 
unsupported features.  The idea is to add unsupported features 
only for likely paths, as follows: first train a CRF without any 
unsupported features, stopping after a few iterations; then add 
unsupported features $f_{pk} (\ys_c, \xs_c)$ for cases where $\xs_c$ 
occurs in the training data for some instance $\inst{x}{i}$, and $p(\ys_c | \inst{x}{i}) > \epsilon$. 

\citeauthornum{mccallum03ifcrf} presents a more principled method of feature 
induction for CRFs, in which the model begins with a number of base features,
and the training procedure adds conjunctions of those features.  Alternatively,
one can use feature selection.
A modern method for feature selection is $L_1$ regularization, which we discuss in Section~\ref{sec:lc-crf-estimation}. \citeauthornum{lavergne10crfs} find that in the most favorable cases
$L_1$ finds models in which only 1\% of the full feature set is non-zero, but with comparable 
performance to a dense feature setting.  They also find it useful, after optimizing the $L_1$-regularized likelihood to find a set of nonzero features,
to fine-tune the weights of the nonzero features only using an $L_2$-regularized objective.

Second, if the observations are categorical rather 
than ordinal, that is, if they are discrete but have 
no intrinsic order, it is important to convert them 
to binary features.  For example, it makes sense to 
learn a linear weight on $f_k(y, x_t)$ when $f_k$ is 1 if $x_t$ 
is the word \emph{dog} and 0 otherwise, but not 
when $f_k$ is the integer index of word $x_t$ 
in the text's vocabulary. 
Thus, in text applications, CRF features are typically 
binary; in other application areas, such as vision and  
speech, they are more commonly real-valued. 
For real-valued features, it can help to apply standard tricks such as
normalizing the features to have mean 0 and standard deviation 1 or to
bin the features to convert them to categorical values.

Third, in language applications, it is sometimes helpful to include 
redundant factors in the model.  For example, in a linear-chain CRF, 
one may choose to include both edge factors $\Psi_t(y_t, y_{t-1}, 
\xs_t)$ and variable factors $\Psi_t (y_t, \xs_t)$.  Although one 
could define the same family of distributions using only edge factors, 
the redundant node factors provide a kind of backoff, which is useful 
when the amount of data is small compared to the number of features.
(When there are hundreds of thousands of features, many data sets are small!)
 It is important to use regularization (Section~\ref{sec:lc-crf-estimation}) when using redundant 
features because it is the penalty on large weights 
that encourages the weight to be spread across the overlapping 
features.

\section{Notes on Terminology} 
 
Different parts of the theory of graphical models have been developed 
independently in many different areas, so many of the concepts in this 
chapter have different names in different areas.  For example, 
undirected models are commonly also referred to \emph{Markov random 
  fields}, \emph{Markov networks}, and \emph{Gibbs distributions}.  As 
mentioned, we reserve the term ``graphical model'' for a family of 
distributions defined by a graph structure; ``random field'' or 
``distribution'' for a single probability distribution; and 
``network'' as a term for the graph structure itself.  This choice of terminology is not always consistent in the literature, partly because it is not ordinarily necessary to be precise in separating these concepts. 
 
Similarly, directed graphical models are commonly known as 
\emph{Bayesian networks}, but we have avoided this term because of its 
confusion with the area of Bayesian statistics.  The term 
\emph{generative model} is an important one that is commonly used in 
the literature, but is not usually given a precise definition.

 

\chapter{Inference}
\label{chp:inference}

Efficient inference is critical for CRFs, both during training and for
predicting the labels on new inputs.
The are two inference problems that arise.  First, after we have
trained the model, we often predict the labels of a new input $\bx$
using the most likely labeling $\ys^* = \arg
\max_{\ys} p(\ys|\xs)$.  Second, as will be seen in
Chapter~\ref{chp:training}, estimation of the parameters typically
requires that we compute the marginal distribution for each edge
$p(y_t, y_{t-1}|\xs)$, and also the normalizing function $Z(\xs)$.

These two inference problems can be seen as fundamentally the same operation on two different 
semirings \citep{aji00gdl}, that is, to change the marginalization problem to the maximization problem, 
we simply substitute max for plus.  Although for discrete variables the marginals can be computed by brute-force summation, 
the time required to do this is exponential 
in the size of $Y$. 
Indeed, both inference problems are intractable for general graphs, because 
any propositional satisfiability problem can be easily represented as 
a factor graph. 

In the case of linear-chain CRFs, both inference tasks can be performed efficiently 
and exactly by variants of the standard dynamic-programming algorithms 
for HMMs.  We begin by presenting these algorithms---the
forward-backward algorithm for computing marginal distributions and 
Viterbi algorithm for computing the most probable assignment---in
Section~\ref{sm:sec:lc-inference}. 
 These algorithms are a
special case of the more general belief propagation algorithm for
tree-structured graphical models
(Section~\ref{sec:belief-propagation}).
For more complex models, approximate inference is necessary.
In principle, we could run any approximate inference algorithm we
want, and substitute the resulting approximate marginals for the exact
marginals within the gradient \eqref{sm:eqn:gradient1}.
This can cause issues, however, because for many optimization
procedures, such as BFGS, we require an approximation to the
likelihood function as well.
We discuss this issue in Section~\ref{sec:alternative-training}.

In one sense, the inference problem for a CRF is no different than that for any graphical
model, so any inference algorithm for graphical models can be used, as
described in several textbooks
\cite{mackay:book,koller:book}.  However, there are two additional issues that need to
be kept in mind in the context of CRFs. 
The first issue is that the inference subroutine is
called repeatedly during parameter estimation (Section~\ref{sec:lc-crf-estimation} explains why), which can be
computationally expensive, so we may wish to trade off inference
accuracy for computational efficiency.  
The second issue is that when approximate inference is used, there can
be complex interactions between the inference procedure and the
parameter estimation procedure. 
 We postpone discussion of
these issues to Chapter~\ref{chp:training}, when we discuss parameter
estimation, but it is worth mentioning
them here because they strongly influence the choice of inference
algorithm.

\section{Linear-Chain CRFs} 
\label{sm:sec:lc-inference} 
 
 In this section, we briefly 
review the inference algorithms for HMMs, the forward-backward and
Viterbi algorithms, and describe how they can be applied to linear-chain CRFs. 
These standard inference algorithms are described in more detail by 
\citeauthornum{rabiner89hmm}.  Both of these algorithms are special cases of the 
belief propagation algorithm described in Section~\ref{sec:belief-propagation},  
but we discuss the special case of linear chains in detail both because it may help to make the earlier discussion  
more concrete, and because it is useful in practice. 
 
First, we introduce notation which will simplify the forward-backward recursions.  An HMM can be viewed as a factor 
graph $p(\ys,\xs) = \prod_t \Psi_t (y_t, y_{t-1}, x_t)$ where 
$Z=1$, and the factors are defined as: 
\begin{equation} 
\Psi_t (j,i,x) \defas p(y_t = j | y_{t-1} = i) p(x_t = x | y_t = j). 
\end{equation} 
If the HMM is viewed as a weighted finite state machine, then  
$\Psi_t (j,i,x)$ is the weight on the transition from state $i$ to state $j$ when the current observation is $x$.  
 
Now, we review the HMM forward algorithm, which is used to compute the probability $p(\xs)$ of the observations.  The idea behind forward-backward is to first rewrite the naive summation $p(\xs) = \sum_{\ys} p(\xs,\ys)$ using the distributive law: 
\begin{align} 
p(\xs) &= \sum_{\ys} \prod_{t=1}^T \Psi_t(y_t, y_{t-1}, x_t) \\ 
           &= \sum_{y_\sT} \sum_{y_{\sT-1}} \Psi_\sT (y_\sT,  y_{\sT-1}, x_\sT) \sum_{y_{\sT-2}} \Psi_{\sT-1} (y_{\sT-1}, y_{\sT-2}, x_{\sT-1}) 
\sum_{y_{\sT-3}} \cdots  
         \label{sm:eqn:fwd-intuition} 
\end{align} 
Now we observe that each of the intermediate sums is reused many times during the computation of the outer sum, and so we can save an exponential amount of work by caching the inner sums. 
 
This leads to defining a set of \emph{forward variables} $\alpha_t$, each of which is a vector of size $M$ (where $M$ is the number of states) 
which stores one of the intermediate sums.  These are defined as: 
\begin{align} 
\alpha_t (j) &\defas p(\xs_\range{1}{t}, y_t = j) \\ 
  &\,= \sum_{\ys_\range{1}{t-1}} \Psi_t (j, y_{t-1}, x_t)  \prod_{t' = 1}^{t-1} \Psi_{t'} (y_{t'}, y_{t'-1}, x_{t'}), 
\end{align} 
where the summation over $\ys_\range{1}{t-1}$ ranges over all assignments to the sequence of random variables $y_1, y_2, \ldots, y_{t-1}$. 
The alpha values can be computed by the recursion 
\begin{equation} 
\alpha_t (j) = \sum_{i \in S} \Psi_t(j,i,x_t) \alpha_{t-1}(i),\label{sm:eqn:hmm-fwd} 
\end{equation} 
with initialization $\alpha_1 (j) = \Psi_1 (j, y_0, x_1)$.  (Recall that $y_0$ is the fixed initial state of the HMM.) 
It is easy to see that $p(\xs) = \sum_{y_\sT} \alpha_\sT (y_\sT)$ by repeatedly substituting the recursion \eq{sm:eqn:hmm-fwd} to obtain \eq{sm:eqn:fwd-intuition}.  A formal proof would use induction.   
 
The backward recursion is exactly the same, except that in \eq{sm:eqn:fwd-intuition}, we push in the summations in reverse order.   
This results in the definition 
\begin{align} 
\beta_t (i) &\defas p(\xs_{\range{t+1}{\sT}} | y_t = i) \\ 
 &= \sum_{\ys_\range{t+1}{\sT}} \prod_{t'=t+1}^{\sT} \Psi_{t'} (y_{t'}, y_{t'-1}, x_{t'}), 
\end{align} 
and the recursion 
\begin{equation} 
\beta_t (i) = \sum_{j \in S} \Psi_{t+1} (j,i,x_{t+1}) \beta_{t+1}(j),\label{sm:eqn:hmm-bwd} 
\end{equation} 
which is initialized $\beta_\sT (i) = 1$. 
Analogously to the forward case, we can compute $p(\xs)$ 
using the backward variables as 
$p(\xs) = \beta_0 (y_0) \defas \sum_{y_1} \Psi_1 (y_1, y_0, x_1) \beta_1(y_1)$. 
 
By combining results from the forward and backward recursions, we can 
compute the marginal distributions $p(y_{t-1}, y_t | \xs)$ needed for the gradient 
\eq{sm:eqn:lc-lik-derivative}.  This can be seen from either the 
probabilistic or the factorization perspectives.  First, taking a 
probabilistic viewpoint we can write 
\begin{align} 
p(y_{t-1}, y_t | \xs) &= \frac{p(\xs | y_{t-1}, y_t) p(y_t, y_{t-1})}{p(\bx)}  \\ 
   &= \frac{p(\xs_{\range{1}{t-1}}, y_{t-1}) p(y_{t} | y_{t-1}) p(x_t | y_t) p(\xs_{\range{t+1}{\sT}} | y_t)}{p(\bx)} \\ &\propto \alpha_{t-1} (y_{t-1}) \Psi_t (y_t, y_{t-1}, x_t) \beta_t (y_t), 
\end{align} 
where in the second line we have used the fact that $\xs_{\range{1}{t-1}}$ is independent from $\xs_{\range{t+1}\sT}$ and from $x_t$ given $y_{t-1}, y_t$. 
Equivalently, from the factorization perspective, we can apply the distributive law to obtain  
we see that 
\begin{multline} 
p (y_{t-1}, y_t, \xs) = \Psi_{t} (y_{t}, y_{t-1}, x_t) \\ \left( \sum_{\ys_\range{1}{t-2}} \prod_{t' = 1}^{t-1} \Psi_{t'} (y_{t'}, y_{t'-1}, x_{t'}) \right) \\ \left(\sum_{\ys_\range{t+1}{\sT}} \prod_{t' = t+1}^T \Psi_{t'} (y_{t'}, y_{t'-1}, x_{t'}) \right), 
\end{multline} 
which can be computed from the forward and backward recursions as 
\begin{equation} 
p (y_{t-1}, y_t, \xs) = \alpha_{t-1} (y_{t-1}) \Psi_{t} (y_{t}, y_{t-1} ,x_t) \beta_t (y_t). 
\end{equation} 
Once we have $p(y_{t-1}, y_t, \xs)$, we can renormalize over $y_t, y_{t-1}$ to obtain the desired marginal $p(y_{t-1}, y_t | \xs)$. 
 
Finally, to compute the globally most probable assignment $\ys^* = \arg \max_{\ys} p(\ys|\xs)$, we observe that the trick in \eq{sm:eqn:fwd-intuition} still works if all the summations are replaced by maximization.  This yields the Viterbi recursion: 
\begin{equation} 
\delta_t (j) = \max_{i \in S} \Psi_t (j, i, x_t) \delta_{t-1}(i)\label{sm:eqn:hmm-viterbi} 
\end{equation} 
 
Now that we have described the forward-backward and Viterbi 
algorithms for HMMs, the generalization to linear-chain CRFs 
is fairly straightforward. 
The forward-backward algorithm for linear-chain CRFs is 
 identical to the HMM version, except that the transition weights $\Psi_t (j,i,x_t)$ are defined differently.  We observe that the CRF model \eq{sm:eqn:lc-crf} can be rewritten as: 
\begin{equation} 
p(\ys|\xs) = \frac{1}{Z(\xs)} \prod_{t=1}^T \Psi_t (y_{t}, y_{t-1}, \xs_t), 
\end{equation} 
where we define 
\begin{equation} 
\Psi_t (y_{t}, y_{t-1}, \xs_t) = \exp \left\{ \sum_k \theta_k f_k (y_{t}, y_{t-1}, \xs_t) \right\}. \label{crf:psi}
\end{equation} 
 
With that definition, the forward recursion \eq{sm:eqn:hmm-fwd}, the backward recursion \eq{sm:eqn:hmm-bwd}, and the Viterbi recursion \eq{sm:eqn:hmm-viterbi} can be used unchanged for linear-chain CRFs. 
Instead of computing $p(\xs)$ as in an HMM, in a CRF  
the forward and backward recursions compute $Z(\xs)$. 

We mention three more specialised inference tasks that can also be
solved using direct analogues of the HMM algorithms.
First, assignments to $\by$ can be sampled from the joint
posterior $p(\by | \bx)$ using the forward algorithm combined with a
backward sampling place, in exactly the same way as an HMM.
Second, if instead of finding the single best assignment
$\arg\max_{\by} p(\by | \bx)$, we wish to find the $k$ assignments
with highest probability,
we can do this also using the standard
algorithms from HMMs.
Finally, sometimes it is useful 
to compute a marginal probability $p(y_t, y_{t+1}, \ldots y_{t+k} | \xs)$ over a possibly non-contiguous range of nodes.  For example, this is useful 
for measuring the model's confidence in its predicted labeling 
over a segment of input.  This marginal probability can be computed 
efficiently using constrained forward-backward, 
as described by \citeauthornum{culotta04confidence}. 

\section{Inference in Graphical Models} 
 
Exact inference algorithms for general graphs exist.  Although these algorithms require exponential time in the worst case, they can still be efficient for graphs that occur in practice.  The most popular exact
algorithm, the junction tree algorithm, successively clusters 
variables until the graph becomes a tree.  Once an equivalent tree has been 
constructed, 
its marginals can be computed using 
exact inference algorithms that are specific to trees.
  However, for certain complex graphs, the junction tree 
algorithm is forced to make clusters which are very large, which is why
the procedure still
requires exponential time in the worst case.  For more details on exact inference, see \citeauthornum{koller:book}.
 
For this reason, an enormous amount of effort has been devoted to 
approximate inference algorithms.  Two classes of approximate 
inference algorithms have received the most attention: 
{Monte Carlo} algorithms and variational algorithms.
\emph{Monte Carlo} algorithms are stochastic algorithms that attempt
to approximately produce a sample from the distribution of interest.
\emph{Variational} algorithms are algorithms that convert the
inference problem into an optimization problem, by attempting to find
a simple distribution that most closely matches the intractable
distribution of interest.  Generally, 
Monte Carlo algorithms are unbiased in the sense that they guaranteed to sample from the distribution 
of interest given enough computation time, although it is usually impossible 
in practice to know when that point has been reached.  Variational 
algorithms, on the other hand, can be much faster, but they tend to be 
biased, by which we mean that they tend to have a 
source of error that is inherent to the approximation, and cannot be 
easily lessened by giving them more computation time. 
Despite this, variational algorithms can be useful for CRFs, because parameter estimation requires performing inference 
many times, and so a fast inference procedure is vital to efficient training. 
 
In the remainder of this section, we outline two examples of
approximate inference algorithms, one from each of these two
categories.  Too much work has been done on
approximate inference for us to attempt to summarize it here.  Rather,
our aim is to 
highlight the general issues that arise when using approximate
inference algorithms within CRF training.  In this chapter, we focus on
describing the inference algorithms themselves, whereas in Chapter~\ref{chp:training}
we discuss their application to CRFs.

\subsection{Markov Chain Monte Carlo}
\label{sec:mcmc}

Currently the most popular type of Monte Carlo method for complex models is
\emph{Markov Chain Monte Carlo} (MCMC) \cite{robert:mcmc}.
Rather than attempting to approximate a marginal distribution $p(y_s |
\bx)$ directly, MCMC methods generate approximate samples from the
joint distribution $p(\by | \bx)$.
MCMC methods work by
constructing a Markov chain, whose state space is the same as that of
$Y$, in careful way so that when the chain is simulated for a long time, the
distribution over states of the chain is
approximately $p(y_s | \bx)$.  Suppose that we want to
approximate the expectation of some function $f(\bx, \by)$ that
depends on.   Given a sample $\by^1,
\by^2, \ldots, \by^M$ from a Markov chain in an MCMC method, we can
approximate this expectation as:
\begin{equation}
 \sum_{\by} p(\by | \bx) f(\bx, \by) \approx \frac{1}{M}  \sum_{j=1}^M f(\bx, \by^j)\label{eq:mcmc}
\end{equation}
For example, in the context of CRFs, these approximate expectations can then be used to approximate the
quantities required for learning, specifically the gradient
\eqref{sm:eqn:lc-lik-derivative}.

A simple example of an MCMC method is Gibbs sampling.
In each iteration of the Gibbs sampling algorithm, each variable is
resampled individually, keeping all of the other variables fixed.
Suppose that we already have a sample $\by^j$ from iteration $j$.
Then to generate the next sample $\by^{j+1}$, 
\begin{enumerate}
\item Set $\by^{j+1} \gets \by^j$.
\item For each $s \in V$, resample component $s$.  Sample $\by^{j+1}_s$
  from the distribution $p(y_s | \by_{\backslash s}, \bx)$.
\item Return the resulting value of $\by^{j+1}$.
\end{enumerate}
This procedure defines a Markov chain that can be used to
approximation expectations as in \eqref{eq:mcmc}.
In the case of general CRFs, then using the notation from Section~\ref{sec:mdl:crf-general}, this
conditional probability can be computed as
\begin{equation}
p(y_s | \by_{\backslash s}, \bx) = \kappa \prod_{C_p \in \smCalC} \prod_{\Psi_c \in C_p} \Psi_c (\xs_c, \ys_c; \theta_p),
\end{equation}
where $\kappa$ is a normalizing constant.
This is much easier to compute than the joint probability $p(\by |
\bx)$, because computing $\kappa$ requires a summation only over all
possible values of 
$y_s$ rather than assignments to the full vector $\by$.

A major advantage of Gibbs sampling is that it is simple to implement.
Indeed, software packages such as BUGS can take a graphical model as
input and automatically compile an appropriate Gibbs sampler \cite{lunn00winbugs}.
The main disadvantage of Gibbs sampling is that it can work poorly if
$p(\by | \bx)$ has strong dependencies, which is often the case in
sequential data.
By ``works poorly'' we mean that it may take many iterations before
the distribution over samples from the Markov chain is close to the
desired distribution $p(\by | \bx)$.

There is an enormous literature on MCMC algorithms.
The textbook by \citeauthornum{robert:mcmc} provides an overview.
However, MCMC algorithms are not commonly applied in the context of conditional
random fields.
  Perhaps the main reason for this is that as we have
mentioned earlier, parameter estimation by maximum likelihood requires
calculating marginals many times.  In the most straightforward
approach, one MCMC chain would be run for each training example for
each parameter setting that is visited in the course of a gradient
descent algorithm.  Since MCMC chains can take
thousands of iterations to converge, this can be computationally prohibitive.
One can imagine ways of addressing this, such as not running the chain
all the way to convergence (see Section~\ref{est:sec:mcmc}).

\subsection{Belief Propagation} 
\label{sec:belief-propagation} 
 
An important variational inference algorithm is \emph{belief propagation (BP)}, 
which we explain in this section. In addition, it is a 
direct generalization of the exact inference algorithms for 
linear-chain CRFs.

Suppose that $G$ is a tree, and we wish to compute the marginal 
distribution of a variable $s$.  The intuition behind BP is that each 
of the neighboring factors of $s$ makes a multiplicative contribution to the marginal 
of $s$, called a \emph{message}, and each of these messages can be 
computed separately because the graph is a tree.  More formally, for 
every factor $a \in N(s)$, call $V_a$ the set of variables that are 
``upstream'' of $a$, that is, the set of variables $v$ for which $a$ 
is between $s$ and $v$.  In a similar fashion, call $F_a$ the set of 
factors that are upstream of $a$, including $a$ itself.  But now 
because $G$ is a tree, the sets $\{ V_a \} \cup \{ s \}$ form a 
partition of the variables in $G$. This means that we can split up the 
summation required for the marginal into a product of independent 
subproblems as: 
\begin{align} 
p(y_s) &\propto \sum_{\by \backslash y_s} \prod_a \Psi_a(\by_a) \\ 
       &= \prod_{a \in N(s)} \sum_{\by_{V_a}} \prod_{\Psi_b \in F_a} \Psi_b (\by_b)\label{eq:bk:bp1} 
\end{align} 
Denote each factor in the above equation by $m_{as}$,  that is, 
\begin{equation} 
m_{as}(x_s) = \sum_{\by_{V_a}} \prod_{\Psi_b \in F_a} \Psi_b (\by_b),\label{eq:sm:message-naive} 
\end{equation} 
can be thought of as a \emph{message} from the factor $a$ to the 
variable $s$ that summarizes the impact of the network upstream of $a$ 
on the belief in $s$.  In a similar fashion, we can define messages 
from variables to factors as 
\begin{equation} 
m_{sA} (x_s) = \sum_{\by_{V_s}} \prod_{\Psi_b \in F_s} \Psi_b (\by_b). 
\end{equation} 
Then, from \eq{eq:bk:bp1}, we have that the marginal $p(y_s)$ is 
proportional to the product of all the incoming messages to variable $s$. 
Similarly, factor marginals can be computed as 
\begin{equation} 
p(\by_a) \propto \Psi_a (\by_a) \prod_{s \in a} m_{sa} (\by_a). 
\end{equation} 
Here we treat $a$ as a set a variables denoting the scope of factor $\Psi_{a}$, as we will throughout.  In addition, we will sometimes use the reverse notation $c \ni s$ to mean the set of all factors $c$ that contain the variable $s$. 
 
Naively computing the messages according to~\eq{eq:sm:message-naive} is impractical, because the messages as we have defined them require summation over possibly many variables in the graph. 
Fortunately, the messages can also be written using a recursion that requires only local summation.  The recursion is 
\begin{equation} 
\label{bk:eq:bp} 
\begin{split} 
m_{as} (x_s) &= \sum_{\by_a \backslash y_s} \Psi_a (\by_a) \prod_{t \in a \backslash s} m_{ta} (x_t) \\ 
m_{sa} (x_s) &= \prod_{b \in N(s) \backslash a} m_{bs} (x_s) 
\end{split} 
\end{equation} 
That this recursion matches the explicit definition of $m$ can be seen by repeated substitution, and proven by induction. 
In a tree, it is possible to schedule these recursions such that the antecedent messages are always sent before their dependents, by first sending messages from the root, and so on. 
This is the algorithm known as \emph{belief propagation} \citep{pearl88priis}.   
 
In addition to computing single-variable marginals, we will also wish to compute 
factor marginals  $p(\by_a)$ and joint probabilites 
$p(\by)$ for a given assignment $\by$.  (Recall that the latter problem is difficult because it requires 
computing the partition function $\log Z$.)  First, to compute marginals over factors---or over any connected set of variables, in fact---we can use the same decomposition of the marginal as for the single-variable case, and get 
\begin{equation} 
p(\by_a) = \kappa \Psi_a (\by_a) \prod_{s \in a} m_{sa} (y_s), 
\end{equation} 
where $\kappa$ is a normalization constant.
In fact, a similar idea works for any connected set of variables---not just a set that happens to be the domain of some factor---although if the set is too large, 
then computing $\kappa$ is impractical.   
 
BP can also be used to compute the normalizing constant $Z(\bx)$.
This can be done directly from the propagation algorithm, in an
analogous way to the forward-backward algorithm in Section~\ref{sm:sec:lc-inference}.
Alternatively, there is another way to compute $Z(x)$ from only the
beliefs at the end of the algorithm.  In a tree structured
distribution, it is always true that
\begin{equation} 
p(\by) = \prod_{s \in V} p(y_s) \prod_{a} \frac{p(\by_a)}{\prod_{t \in a} p(y_t)} \label{bk:eq:treejoint} 
\end{equation} 
For example, in a linear chain this amounts to
\begin{equation}
p(\by) = \prod_{t=1}^T p(y_t) \prod_{t=1}^{\sT} \frac{p(y_t, y_{t-1})}{p(y_t) p(y_{t-1})},
\end{equation}
which, after cancelling and rearranging terms, is just another way to
write the familiar equation $p(\by) = \prod_{t} p(y_t | y_{t-1})$.
More generally, \eqref{bk:eq:treejoint} can be derived using the
junction tree theorem, by considering a junction tree with one cluster for each factor. 
Using this identity, we can compute $p(\by)$ (or $\log Z$) from the per-variable  
and per-factor marginals. 

\begin{figure}
  \centering
  \resizebox{3in}{!}{\includegraphics{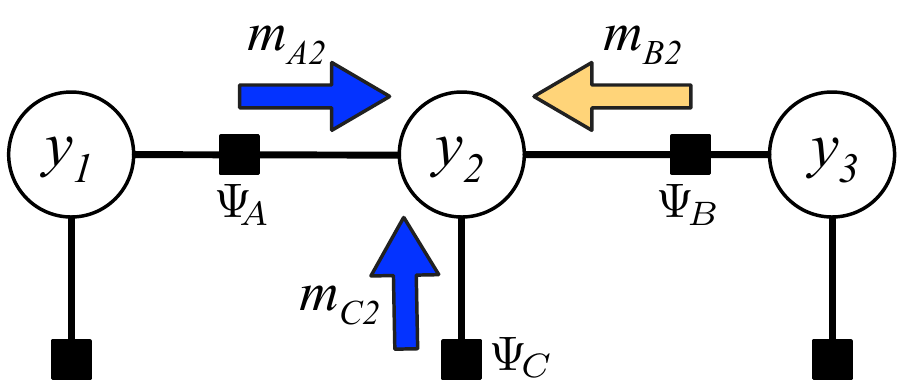}}
  \caption{Illustration of the correspondence between forward backward and belief propagation in linear chain graphs}
  \label{fig:bpfp}
\end{figure}

If $G$ is a tree, belief propagation computes the marginal
distributions exactly.  Indeed, if $G$ is a linear chain, then BP
reduces to the forward-backward algorithm
(Section~\ref{sm:sec:lc-inference}).
To see this, refer to Figure~\ref{fig:bpfp}. 
The figure shows a three node linear chain along with the BP messages
as we have described them in this section.  To see the correspondence
to forward backward, 
the forward message that we denoted $\alpha_2$ in
Section~\ref{sm:sec:lc-inference} corresponds to the product of the
two messages $m_{A2}$ and $m_{C2}$ (the thick, dark blue arrows in
the figure).  The backward message $\beta_2$ corresponds to the
message $m_{B2}$ (the thick, light orange arrow in the figure).
 
If $G$ 
is not a tree, the message updates \eq{bk:eq:bp} are no longer 
guaranteed to return the exact marginals, nor are they guaranteed even 
to converge, but we can still iterate them in an attempt to find a 
fixed point.  This procedure is called \emph{loopy belief 
  propagation}.  To emphasize the approximate nature of this 
procedure, we refer to the approximate marginals that result from loopy 
BP as \emph{beliefs} rather than as marginals, and denote them by 
$q(y_s)$. 
 
Surprisingly, loopy BP can be seen as a variational method for
inference, meaning that there actually exists an objective function
over beliefs that is approximately minimized by the iterative BP procedure.
Several introductory papers 
\cite{yedidia05gbp,wainwright08graphical} describe this in more
detail.

The general idea behind a variational algorithm is: 
\begin{enumerate} 
\item Define a family 
of tractable distributions $\varQFam$ and an objective function 
$\varO(q)$.  The function $\varO$ should be designed to measure how well a tractable distribution $q \in 
\varQFam$ approximates the distribution $p$ of interest.   
\item Find 
the ``closest'' tractable distribution $q^* = \min_{q \in \varQFam} 
\varO(q)$. 
\item Use the marginals of $q^*$ to approximate those of $p$. 
\end{enumerate} 
For example, suppose that we take $\varQFam$ be the set of all possible distributions over $\by$, 
and we choose the objective function
\begin{align} 
\varO(q) &= \KL{q}{p} - \log Z \\ 
        &= -H(q) - \sum_a q(\by_a) \log \Psi_a (\by_a).\label{bk:eq:varkl} 
\end{align} 
Then the solution to this variational problem is $q^* = p$ with 
optimal value $\varO(q^*) = \log Z$.  Solving this particular 
variational formulation is thus equivalent to performing exact 
inference.  Approximate inference techniques can be devised by 
changing the set $\varQFam$---for example, by requiring $q$ to be 
fully factorized---or by using a different objective $\varO$. 
For example, the mean field method arises by requiring $q$ to be fully
factorized, i.e., $q(\by) = \prod_s q_s(y_s)$ for some choice for
$q_s$, and finding the factorized $q$ that most closely matches $p$.
 
With that background on variational methods, let us see how belief 
propagation can be understood in this framework.  We make two 
approximations.  First, we approximate the entropy term $H(q)$ 
of~\eq{bk:eq:varkl}, which as it stands is difficult to compute. If $q$ 
were a tree-structured distribution, then its entropy could be written 
exactly as 
\begin{equation} 
H_\bethe(q) = \sum_a q(\by_a) \log q(\by_a) + \sum_i (1 - d_i) q(y_i) \log q(y_i). 
\end{equation} 
This follows from substituting the junction-tree formulation 
\eq{bk:eq:treejoint} of the joint into the definition of entropy.   If $q$ is not a 
tree, then we can still take $H_\bethe$ as an approximation to $H$ to compute  
the exact variational objective $\varO$. 
This yields the \emph{Bethe free energy}: 
\begin{equation} 
\varO_\bethe(q) = H_\bethe(q) -  \sum_a q(\by_a) \log \Psi_a (\by_a) \label{bk:eq:betheprimal} 
\end{equation} 
The objective $\varO_\bethe$ depends on $q$ only through its 
marginals, so rather than optimizing it over all probability 
distributions $q$, we can optimize over the space of all marginal vectors. 
Specifically, every distribution $q$ has an associated \emph{belief vector} $\bq$, 
 with elements $q_{a;y_a}$ 
for each factor $a$ and assignment $y_a$, and elements 
 $q_{i;y_i}$ for each variable $i$ and assignment $y_i$.   
The space of all possible belief vectors has been 
called the \emph{marginal polytope} \citep{wainwright03polytope}. 
However, for intractable models, the marginal polytope can have extremely complex structure. 
 
This leads us to the second variational approximation made by loopy BP, namely that the objective $\varO_\bethe$ is optimized instead over a relaxation of the marginal polytope.  The relaxation is to require that the beliefs be only \emph{locally consistent}, that is, that 
\begin{align} 
\sum_{\by_a \backslash y_i } q_a(\by_a) = q_i(y_i) \qquad \forall a, i \in a 
\end{align} 
Under these constraints, \citeauthornum{yedidia04gbp} show that constrained 
stationary points of $\varO_\bethe$ fixed points of loopy BP.  So we 
can view the Bethe energy $\varO_\bethe$ as an objective function that 
the loopy BP fixed-point operations attempt to optimize. 
 
This variational perspective provides new insight into the method
that would not be available if we thought of it solely from the
message passing perspective.  One of the most important insights
is that it shows how to use loopy BP to approximate $\log Z$.
Because we introduced $\min_q \varO_\bethe(q)$ as an approximation to $\min_q
\varO(q)$, and we know that $\min_q \varO(q) = \log Z$, then it seems
reasonable to define $\log Z_\bethe = \min_q \varO_\bethe(q)$ as an
approximation to $\log Z$.  This will be important when we discuss
CRF parameter estimation using BP in Section~\ref{est:sec:bp}.

\section{Implementation Concerns}
\label{inf:sec:implementation}

In this section, we mention a few implementation techniques that are important
to practical inference in CRFs: sparsity and preventing numerical underflow.

First, it is often possible to exploit \emph{sparsity} in the model to
make inference more efficient. Two different types of sparsity are
relevant: sparsity in the factor values, and sparsity in the
features.  First, about the factor values, recall that in the
linear-chain case, each of the
forward updates \eqref{sm:eqn:hmm-fwd} and backward updates
\eqref{sm:eqn:hmm-bwd} requires $O(M^2)$ time, that is, quadratic time
in the number of labels.  Analogously, in general CRFs, an update of
loopy BP in a model with pairwise factors requires $O(M^2)$ time.
In some models, however, it is possible to implement inference more
efficiently, because it is known a priori not all factor values $(y_t,
y_{t-1})$ are feasible, that is, the factor $\Psi_t (y_t, y_{t+1},
\bx_t)$ is 0 for many values $y_t, y_{t+1}$.  In such cases, the computational cost of sending a
message can be reduced by implementing the message-passing iterations using
sparse matrix operations.

The second kind of sparsity that is useful is sparsity in the
feature vectors.  Recall from \eqref{eq:crf-factors} that computing
the factors  $\Psi_c (\bx_c, \by_c)$
 requires computing a dot product between the parameter vector
$\theta_p$ and
and the vector of features $F_c = \{ f_{pk}(y_c, \bx_c) \}$.   Often, many
elements of the vectors $F_c$ are zero.  For example, natural
language applications often involve binary indicator variables on word
identity.  In this case, the time required to compute the factors $\Psi_c$ can
be greatly improved using a sparse vector representation.  In a
similar fashion, we can use sparsity  improve the time required to
compute
the likelihood gradient, as we discuss in
Chapter~\ref{chp:training}.
 
A related trick, that will also speed up forward backward, is to tie 
the parameters for certain subsets of transitions \cite{cohn06ecml}.
This has the effect of reducing the effective size of the model's transition matrix,
lessening the effect of the quadratic dependence of the size of the label set.

A second implementation concern that arises in inference is avoiding
numerical underflow.
The probabilities involved in forward-backward and 
belief propagation are often too small to be represented within numerical 
precision (for example, in an HMM they decay toward 0 exponentially fast in $T$).  There are two standard approaches to this common problem. 
One approach is to scale each of the vectors $\alpha_t$ and 
$\beta_t$ to sum to 1, thereby magnifying small values.  This scaling does 
not affect our ability to compute $Z(\bx)$ because it can be computed as $Z(\bx) = p(\by' | \bx)^{-1} \prod_t (\Psi_t(y'_t, y'_{t+1}, \bx_t))$
for an arbitrary assignment $\by'$, where $p(\by' | \bx)^{-1} $ is computed from the marginals using
\eqref{bk:eq:treejoint}. But in fact, there is actually a more efficient method
described by \citeauthornum{rabiner89hmm} that involves saving each of the local scaling factors.  In any case, the scaling trick can be
used in forward-backward or loopy BP; in either case, it does not
affect the final values of the beliefs.
 
A second approach to preventing underflow is to perform computations in the logarithmic domain, e.g., the forward recursion \eqref{sm:eqn:hmm-fwd}  becomes 
\begin{equation} 
\log \alpha_t (j) = \bigoplus_{i \in S} \big(\log \Psi_t(j,i,x_t) + \log \alpha_{t-1}(i) \big),\label{sm:eqn:log-hmm-fwd} 
\end{equation} 
where $\oplus$ is the operator $a \oplus b = \log (e^a + e^b)$. 
At first, this does not seem much of an improvement, since numerical precision is lost when computing $e^a$ and $e^b$.  But $\oplus$ can be computed as 
\begin{equation} 
a \oplus b = a + \log (1 + e^{b-a}) = b + \log (1 + e^{a-b}), 
\end{equation} 
which can be much more numerically stable,  
particularly if we pick the version of the identity with the smaller
exponent.   

At first, it would seem that the normalization approach is preferable to the
logarithmic approach, because the logarithmic approach requires
$O(TM^2)$ calls to the special functions log and exp, which can
be computationally expensive.   This observation is correct for HMMs,
but not for CRFs.
In a CRF, even when the normalization
approach is used, it is still necessary to call the $\exp$ function in
order to compute $\Psi_t(y_t, y_{t+1}, \bx_t)$, defined in
\eqref{crf:psi}.  
So in CRFs, special functions cannot be avoided.  In the
worst case, there are $TM^2$ of these $\Psi_t$ values, so the
normalization approach needs $TM^2$ calls to special functions just as
the logarithmic domain approach does.
However, there are some special cases in which the normalization
approach can yield a speedup, such as when the transition
features do not depend on the observations, so that there are only
$M^2$ distinct $\Psi_t$ values.

\chapter{Parameter Estimation} 
\label{chp:training}

In this chapter we discuss how to estimate the parameters 
$\theta = \{ \theta_k \}$ of a conditional random field.
In the simplest and typical case, we are provided with fully labeled
independent data, but there has also been work in
CRFs with latent variables and CRFs for relational learning.

CRFs are trained by \emph{maximum likelihood}, that is, the parameters
are chosen such that the training data has highest probability under the model.  
In principle, this can be done in a manner exactly analogous
to logistic regression, which should not be surprising given
the close relationship between these models that was described in
Chapter~\ref{chp:graphical}.  The main difference is computational:
CRFs tend to have more parameters and more complex structure than a simple
classifier, so training is correspondingly more expensive.

In tree structured CRFs, the maximum likelihood parameters can be
found by a numerical optimization procedure that calls the
inference algorithms of Section~\ref{sm:sec:lc-inference} as a
subroutine.  Crucially, the likelihood is a convex function of the
parameters, which means that powerful optimization procedures are
available that provably converge to the
optimal solution.
For general CRFs, on the other hand, maximum likelihood training is intractable.  One way
to deal with this problem is to use approximate inference methods,
as discussed in Chapter~\ref{chp:inference}, but another way
is to choose a different training criterion than maximum likelihood.

We begin by describing maximum likelihood training, both in the linear
chain case (Section~\ref{sec:lc-crf-estimation}) and in the case of
general graphical structures (Section~\ref{bk:sec:training-general}),
including the case of latent variables.
Then we discuss training in general graphical structures,  in which
approximations are necessary.  We also describe two general methods
for speeding up parameter estimation that exploit iid structure in the
data: stochastic gradient descent (Section~\ref{est:sec:sg}) and
multithreaded training (Section~\ref{sec:parallel}).
In CRFs with general structure, typically approximate inference
procedures must be used.
The approximate training procedures build on the approximate algorithms for inference described in
Chapter~\ref{chp:inference}, but there can be complications in the
interaction between approximate inference and learning.
This is described in Section~\ref{sec:alternative-training}.

\section{Maximum Likelihood}

\subsection{Linear-chain CRFs}
\label{sec:lc-crf-estimation} 

In a linear-chain CRF, the maximum likelihood parameters can be
determined using numerical optimization methods.
We are given iid training data  $\Data = \{ \inst{\xs}{i},  
\inst{\ys}{i} \}_{i=1}^N$, where each $\inst{\xs}{i} =  \{
\insttvec{x}{i}{1},  \insttvec{x}{i}{2}, \ldots  \insttvec{x}{i}{T}
\}$  is a sequence of inputs, and each $\inst{\ys}{i} = \{
\instt{y}{i}{1},  \instt{y}{i}{2}, \ldots  \instt{y}{i}{T} \}$   is a
sequence of the desired predictions.  
 
Parameter estimation is typically performed by penalized maximum likelihood.  Because we are modeling the conditional 
distribution, the following log likelihood, sometimes called the \emph{conditional log likelihood}, is appropriate: 
\begin{equation} 
\ell(\theta) = \sum_{i=1}^N \log p(\inst{\ys}{i}| \inst{\xs}{i}).\label{sm:eqn:conditional-likelihood} 
\end{equation} 
One way to understand the conditional likelihood $p(\ys|\xs;\theta)$ is to imagine combining it with some arbitrary prior $p(\xs;\theta')$ to form a joint $p(\ys,\xs)$.  Then when we optimize the joint log likelihood  
\begin{equation} 
\log p(\ys,\xs) = \log p(\ys|\xs;\theta) + \log p(\xs;\theta'), 
\end{equation} 
the two terms on the right-hand side are decoupled,  
that is, the value of $\theta'$ does not 
affect the optimization over $\theta$.  If we do not need to 
estimate $p(\xs)$, then we can simply drop the second term,  
which leaves \eq{sm:eqn:conditional-likelihood}. 
 
After substituting in the CRF model \eq{sm:eqn:lc-crf} 
into the likelihood \eq{sm:eqn:conditional-likelihood}, we get the following expression: 
\begin{equation} 
\ell(\theta) = \sum_{i=1}^N \sum_{t=1}^T \sum_{k=1}^K \theta_k f_k (\instt{y}{i}{t}, \instt{y}{i}{t-1}, \instt{\xs}{i}{t}) - \sum_{i=1}^N \log Z(\inst{\xs}{i}),\label{sm:eqn:lc-crf-likelihood} 
\end{equation} 
It is often the case that we have a large number of parameters, e.g.,
several hundred thousand.  As a measure to avoid overfitting, we use \emph{regularization}, which is a penalty on weight vectors whose norm is too large.  A common 
choice of penalty is based on the Euclidean norm of $\theta$ 
and on a  
\emph{regularization parameter} $1/2\sigma^2$ that determines 
the strength of the penalty.  Then the regularized log likelihood is 
\begin{equation} 
\ell(\theta) = \sum_{i=1}^N \sum_{t=1}^T \sum_{k=1}^K \theta_k f_k (\instt{y}{i}{t}, \instt{y}{i}{t-1}, \instt{\xs}{i}{t}) - \sum_{i=1}^N \log Z(\inst{\xs}{i}) - \sum_{k=1}^K \frac{\theta_k^2}{2\sigma^2}.\label{sm:eqn:lc-crf-regularized} 
\end{equation} 
The parameter $\sigma^2$ is a 
free parameter which determines how much to penalize large weights. 
Intuitively, the idea is to reduce the potential for a small number of
features to dominate the prediction.
The notation for the regularizer is intended to suggest that 
regularization can also be viewed as performing maximum a posteriori (MAP)
estimation of $\theta$, if $\theta$ is assigned a Gaussian prior with 
mean 0 and covariance $\sigma^2 I$.   
Determining the best regularization parameter can require a 
computationally-intensive parameter sweep.  Fortunately, often the 
accuracy of the final model is not sensitive to small changes 
in $\sigma^2$ (e.g., up to a factor of $10$).  The best value of $\sigma^2$
depends on the size of the training set; for medium-sized training sets, $\sigma^2 = 10$ is typical.

An alternative choice of regularization is to use the $L_1$ norm 
instead of the Euclidean norm, which corresponds to an exponential 
prior on parameters \citep{goodman04exponential}.  This results in the
following penalized likelihood:
\begin{equation}
\ell'(\theta) = \sum_{i=1}^N \sum_{t=1}^T \sum_{k=1}^K \theta_k f_k
(\instt{y}{i}{t}, \instt{y}{i}{t-1}, \instt{\xs}{i}{t}) - \sum_{i=1}^N
\log Z(\inst{\xs}{i}) - \alpha \sum_{k=1}^K |\theta_k|.\label{sm:eqn:lc-crf-l1}
\end{equation} 
This regularizer 
tends to encourage sparsity in the learned parameters, meaning that
most of the $\theta_k$ are 0.  This can be useful for performing 
feature selection, and also has 
theoretical advantages \citep{ng04l1}. 
In practice, models trained with the $L_1$ regularizer tend to be
sparser but have roughly the same accuracy as models training using
the $L_2$ regularizer \cite{lavergne10crfs}.
A disadvantage of the $L_1$ regularizer is that it is not differentiable at 0, which complicates numerical parameter estimation 
somewhat \citep{goodman04exponential,andrew07scalable,yu10quasi}. 

In general, the function $\ell(\theta)$ cannot be maximized in closed form, so numerical optimization is used.  The partial derivatives of \eq{sm:eqn:lc-crf-regularized} are 
\begin{equation} 
\frac{\partial \ell}{\partial \theta_k} = \sum_{i=1}^N \sum_{t=1}^T  f_k (\instt{y}{i}{t}, \instt{y}{i}{t-1}, \instt{\xs}{i}{t}) -  \sum_{i=1}^N \sum_{t=1}^T \sum_{y,y'} f_k (y,y', \instt{\xs}{i}{t}) p(y,y'|\inst{\xs}{i}) - \frac{\theta_k}{\sigma^2}.\label{sm:eqn:lc-lik-derivative} 
\end{equation} 
The first term is the expected value of $f_k$ under the empirical distribution: 
\begin{equation} 
\tilde{p}(\ys,\xs) = \frac{1}{N} \sum_{i=1}^N \Ind{\ys = \inst{\ys}{i}}\Ind{\xs = \inst{\xs}{i}}. 
\end{equation} 
  The second term, which arises from the derivative of $\log Z(\xs)$, is the expectation of $f_k$ under the model distribution $p(\ys|\xs; \theta) \tilde{p}(\xs)$.  Therefore, at the 
  unregularized maximum likelihood 
  solution, when the gradient is zero, these two expectations  
  are equal.  This pleasing interpretation is a standard result 
  about maximum likelihood estimation in exponential families. 

To compute the likelihood $\ell(\theta)$ and its derivative requires
techniques from inference in graphical models.  In the likelihood,
inference is needed to compute the partition function $Z(\inst{\xs}{i})$, which
is a sum over all possible labellings.  In the derivatives, inference
is required to compute the marginal distributions $p(y,y' |
\inst{\xs}{i})$.   Because both of these quantities depend on
$\inst{\xs}{i}$,  we will need
to run inference once for each training instance  every time the
likelihood is computed.  This is the key computational difference
between CRFs and generative Markov random fields.  In linear-chain
models, inference can be performed efficiently using the algorithms
described in Section~\ref{sm:sec:lc-inference}.

Now we discuss how to optimize $\ell(\theta)$. 
The function $\ell(\theta)$ is concave, which follows from the convexity of functions of the form $g(\xs) = \log \sum_i \exp {x_i}$.  Convexity is extremely helpful for parameter estimation, because it means that every local optimum is also a global optimum. 
   Adding regularization ensures that $\ell$ is strictly concave, which implies that it has exactly one global optimum. 
 
Perhaps the simplest approach to optimize $\ell$ is steepest ascent
along the gradient \eq{sm:eqn:lc-lik-derivative}, but this requires
too many iterations to be practical.   Newton's method converges much
faster because it takes into account the curvature of the likelihood,
but it requires computing the Hessian, the matrix of all second
derivatives. 
The size of the Hessian is quadratic in the number of parameters.
Since practical applications often use tens of thousands or even
millions of parameters, simply storing the full Hessian is not
practical.   

Instead, current techniques for optimizing 
\eq{sm:eqn:lc-crf-regularized} make approximate use of second-order 
information.  Particularly successful have been quasi-Newton methods 
such as BFGS \citep{bertsekas99nonlinear}, which compute an 
approximation to the Hessian from only the first derivative of the 
objective function.  A full $K \times K$ approximation to the Hessian 
still requires quadratic size, however, so a limited-memory version of 
BFGS is used, due to \citeauthornum{byrd94bfgs}.  
Conjugate gradient is another optimization 
technique that also makes approximate use of second-order information 
and has been used successfully with CRFs.  For a good introduction to
both limited-memory BFGS and conjugate gradient, see \citeauthornum{nocedal99numerical}. Either can be thought of as 
a black-box optimization routine that is a drop-in replacement for 
vanilla gradient ascent.  When such second-order methods are used, 
gradient-based optimization is much faster than the original 
approaches based on iterative scaling in \citeauthornum{lafferty01crf}, as 
shown experimentally by several authors 
\citep{sha03shallow,wallach02efficient,malouf02comparison,minka03comparison}. 
Finally, trust region methods have recently been shown to 
perform well on multinomial logistic regression \citep{lin07trust},  
and may work well for CRFs as well. 
 
Finally, we discuss the computational cost of training linear chain
models.  As we will see in Section~\ref{sm:sec:lc-inference}, the
likelihood and gradient for a single training instance can be computed
by forward-backward in time $O(TM^2)$, where $M$ is the number of
labels and $T$ the length of the training instance.  Because we need
to run forward-backward for each training instance, each computation
of the likelihood and gradient requires $O(TM^2N)$ time, so that the
total cost of training is $O(TM^2NG)$, where $G$ the number of
gradient computations required by the optimization procedure.
Unfortunately, $G$ depends on the data set and is difficult to predict
in advance.  For batch L-BFGS on linear-chain CRFs, it is often but
not always under 100.  For many data sets, this cost is reasonable,
but if the number of states $M$ is large, or the number of training
sequences $N$ is very large, then this can become expensive.
Depending on the number of labels, training CRFs can take anywhere
from a few minutes to a few days; see Section~\ref{sec:times} for
examples.

\subsection{General CRFs} 
\label{bk:sec:training-general} 
 
Parameter estimation for general CRFs is essentially the same as for linear-chains, except that computing the model expectations requires more general inference algorithms.   
First, we discuss the fully-observed case, in which the training and
testing data are independent, and the training data is fully observed.
In this case the conditional log likelihood,
using the notation of Section~\ref{sec:mdl:crf-general}, is
\begin{equation} 
\ell(\theta) = \sum_{C_p\in \smCalC} \sum_{\Psi_c \in C_p} \sum_{k=1}^{K(p)}  \theta_{pk} f_{pk} (\xs_c, \ys_c) - \log Z(\xs).  \label{eq:genl-lik}
\end{equation} 
The equations in this section do not  
explicitly sum over training instances, because if a 
particular application happens to have iid training 
instances, they 
can be represented by disconnected components in the graph $G$. 
 
The partial derivative of the log likelihood with respect to a parameter $\theta_{pk}$ associated with a clique template $C_p$ is 
\begin{equation} 
\frac{\partial \ell}{\partial \theta_{pk}} =  
 \sum_{\Psi_c \in C_p} f_{pk} (\xs_c, \ys_c) - \sum_{\Psi_c \in C_p} \sum_{\ys'_c} f_{pk} (\xs_c, \ys'_c) p(\ys'_c | \xs). 
    \label{sm:eqn:gradient1} 
\end{equation} 
The function $\ell(\theta)$ has many of the same properties as in the linear-chain case.  First, the zero-gradient conditions can be interpreted as requiring that the sufficient statistics $F_{pk} (\xs, \ys) = \sum_{\Psi_c} f_{pk} (\xs_c, \ys_c)$ have the same expectations under the empirical distribution and under the model distribution.  Second, the function $\ell(\theta)$ is concave, and can be efficiently maximized by second-order techniques such as conjugate gradient and L-BFGS.  Finally, regularization is used just as in the linear-chain case.   
 
All of the discussion so far has assumed that the training data
contains the true values of all the label variables in the model.
In the latent variable case, on the other hand, the model contains
variables that are observed at neither training nor test time. 
This situation is called a \emph{hidden-state CRF (HCRF)}
by \citeauthornum{quattoni05conditional} which was one of the first examples of
latent variable CRFs.  \citeauthornum{quattoni07hidden} present a more
detailed description.
For other early applications of HCRFs, see \citep{sutton07dcrf,mccallum05stringedit}.
It is more difficult to train 
CRFs with latent variables because the latent variables  
need to be marginalized out to compute the likelihood. 
Because of this difficultly, the original work on CRFs 
focused on fully-observed training data, but 
recently there has been increasing 
interest in HCRFs.
 
Suppose we have a conditional random field with inputs $\xs$ 
in which the output variables $\ys$ are observed in 
the training data, but we have additional variables $\ws$ 
that are latent, so that the CRF has the form 
\begin{equation} 
p(\ys,\ws|\xs) = \frac{1}{Z(\xs)} \prod_{C_p \in \smCalC} \prod_{\Psi_c \in C_p} \Psi_c (\xs_c, \ws_c, \ys_c; \theta_p). 
\label{eqn:latent-crf} 
\end{equation} 
A natural objective function to maximize during 
training is the marginal likelihood 
\begin{equation} 
\ell (\theta) = \log p(\ys | \xs) = \log \sum_{\ws} p(\ys, \ws | \xs). 
\end{equation} 
The first question is how even to compute the marginal likelihood  
$\ell (\theta)$, because if there are many variables $\ws$, the sum 
cannot be computed directly.  The key is to realize that we need to compute $ \log \sum_{\ws} p(\ys, \ws | \xs)$ not 
for any possible assignment $\ys$, but only for the particular  
assignment that occurs in the training data.  This motivates 
taking the original CRF \eq{eqn:latent-crf}, and clamping  
the variables $Y$ to their observed values in the training 
data, yielding a distribution over $\ws$: 
\begin{equation} 
p(\ws | \ys, \xs) = \frac{1}{Z(\ys,\xs)}  \prod_{C_p \in \smCalC} \prod_{\Psi_c \in C_p} \Psi_c (\xs_c, \ws_c, \ys_c; \theta_p), 
\label{eqn:latent-crf-clamped} 
\end{equation} 
where the normalization factor is 
\begin{equation} 
Z(\ys,\xs) = \sum_{\ws} \prod_{C_p \in \smCalC} \prod_{\Psi_c \in C_p} \Psi_c (\xs_c, \ws_c, \ys_c; \theta_p). 
\end{equation} 
This new normalization constant $Z(\ys,\xs)$ can be computed by the same inference algorithm 
that we use to compute $Z(\xs)$.  In fact, $Z(\ys,\xs)$ is easier 
to compute, because it sums only over $\ws$, while $Z(\xs)$ sums 
over both $\ws$ and $\ys$.  Graphically, this amounts to saying 
that clamping the variables 
$\ys$ in the graph $G$ can simplify the structure among $\ws$. 
 
Once we have $Z(\ys,\xs)$, the marginal likelihood can 
be computed as 
\begin{equation} 
p(\ys|\xs) = \frac{1}{Z(\xs)} \sum_\ws \prod_{C_p \in \smCalC} \prod_{\Psi_c \in C_p} \Psi_c (\xs_c, \ws_c, \ys_c; \theta_p) = \frac{Z(\ys,\xs)}{Z(\xs)}. 
\end{equation} 
 
Now that we have a way to compute $\ell$, we discuss how 
to maximize it with respect to $\theta$. 
Maximizing $\ell(\theta)$ can be 
difficult because $\ell$ is no longer convex in general 
(log-sum-exp is convex, but the difference of two log-sum-exp functions might not be),  
so optimization procedures are typically guaranteed 
to find only local maxima.  Whatever optimization technique 
is used, the model parameters must be carefully initialized 
in order to reach a good local maximum. 
 
We discuss two different ways to maximize $\ell$: directly 
using the gradient, as in \citeauthornum{quattoni05conditional}; 
and using EM, as in \citeauthornum{mccallum05stringedit}. 
(In addition, it is also natural to use stochastic gradient 
descent here; see Section~\ref{est:sec:sg}.)
To maximize $\ell$ directly, we need to calculate 
its gradient.  The simplest way to do this is to use 
the following fact.   
For any function $f(\theta)$, we have 
\begin{equation} 
\frac{df}{d\theta} = f(\theta) \frac{d \log f}{d\theta}, 
\end{equation} 
which can be seen by applying the chain rule to $\log f$ 
and rearranging.  Applying this to the marginal likelihood 
$\ell (\theta) = \log \sum_\ws p(\ys, \ws | \xs)$ yields 
\begin{align} 
\frac{\partial \ell}{\partial \theta_{pk}} &= \frac{1}{\sum_\ws p(\ys, \ws | \xs)} \sum_\ws \frac{\partial}{\partial \theta_{pk}} \big[ p(\ys, \ws | \xs) \big] \\ 
  &=  \sum_\ws p(\ws | \ys, \xs) \frac{\partial}{\partial \theta_{pk}} \big[ \log p(\ys, \ws | \xs) \big]. 
\end{align} 
This is the expectation of the fully-observed gradient, where 
the expectation is taken over $\ws$.  This expression 
simplifies to 
\begin{multline} 
\frac{\partial \ell}{\partial \theta_{pk}}  
 = \sum_{\Psi_c \in C_p} \sum_{\ws'_c} p(\ws'_c | \ys, \xs) f_k
 (\ys_c, \xs_c, \ws'_c) \\ - \sum_{\Psi_c \in C_p} \sum_{\ws'_c, \ys'_c} p(\ws'_c, \ys'_c | \xs_c) f_k(\ys'_c, \xs_c, \ws'_c).\label{eqn:latent-gradient} 
\end{multline} 
This gradient requires computing two different  
kinds of marginal probabilities. 
The first term contains a marginal probability   
$p(\ws'_c | \ys, \xs)$, 
which is exactly a marginal distribution  
of the clamped CRF \eq{eqn:latent-crf-clamped}. 
  The second term contains a different marginal $p(\ws'_c, \ys'_c | \xs_c)$, which is the same marginal probability  
  required in a fully-observed CRF. 
Once we have computed the gradient, $\ell$ can 
be maximized by standard techniques such as conjugate gradient. 
For BFGS, it has been our experience that the memory-based approximation to the Hessian can become confused by violations of convexity, such as occur in
latent-variable CRFs.  One practical trick in this situation is to reset the Hessian approximation when that happens.

Alternatively, $\ell$ can be optimized using expectation maximization 
(EM).  At each iteration $j$ in the EM algorithm, 
the current parameter vector $\theta^{(j)}$ is updated as 
follows. 
First, in the E-step, 
an auxiliary function $q(\ws)$ is computed as 
$q(\ws) = p(\ws | \ys, \xs; \theta^{(j)})$. 
Second, in the M-step, a new parameter vector $\theta^{(j+1)}$ 
is chosen as 
\begin{equation} 
\theta^{(j+1)} = \arg \max_{\theta'} \sum_{\ws'} q (\ws') \log p(\ys, \ws' | \xs; \theta' ).\label{eqn:m-step} 
\end{equation}   
The direct maximization algorithm and the EM algorithm 
are strikingly similar.  This can 
be seen by substituting the definition of $q$ into \eq{eqn:m-step} 
and taking derivatives. 
The gradient is almost identical to the direct gradient \eq{eqn:latent-gradient}. 
The only difference is that in EM, the distribution $p(\ws|\ys,\xs)$ 
is obtained from a previous, fixed parameter setting 
rather than from the argument of the maximization.  We are unaware 
of any empirical comparison of EM to direct optimization for 
latent-variable CRFs. 
 
\section{Stochastic Gradient Methods}
\label{est:sec:sg}

So far, all of the methods that we have discussed for optimizing the
likelihood work in a \emph{batch setting}, meaning that they do not
make any change to the model parameters until they have scanned the
entire training set. 
If the training data consist of a large number of iid samples,
then this may seem wasteful.  We may suspect that many different items
in the training data provide similar information about the model
parameters, so that it should be possible to update the parameters
after seeing only a few examples, rather than sweeping through all of
them.

\emph{Stochastic gradient descent} (SGD) is a simple optimization method that is
designed to exploit this insight.  The basic idea is at every iteration, to pick a training
instance at random, and take a small step in the direction given by
the gradient for that instance only.
In the batch setting, gradient descent is generally a poor
optimization method, because the direction of steepest descent locally
(that is, the negative gradient) can point in a very different
direction than the optimum. 
So stochastic gradient methods involve an interesting tradeoff:
the directions of the individual steps may be much better in L-BFGS
than in SGD, but the SGD directions can be computed much faster.

In order to keep the notation simple,
we present SGD only for the
case of linear-chain CRFs, but it can be
easily used with any graphical structure, as long as the training
data are iid.
The gradient of the likelihood for a single training instance
$(\inst{\xs}{i},  \inst{\ys}{i})$ is
\begin{equation}
\frac{\partial \ell_i}{\partial \theta_k} = \sum_{t=1}^T  f_k
(\instt{y}{i}{t}, \instt{y}{i}{t-1}, \instt{\xs}{i}{t}) - \sum_{t=1}^T
\sum_{y,y'} f_k (y,y', \instt{\xs}{i}{t}) p(y,y'|\inst{\xs}{i}) -
\frac{\theta_k}{N \sigma^2}.\label{eq:sgd-gradient}
\end{equation}
This is exactly the same as the full gradient
\eqref{sm:eqn:lc-lik-derivative}, with two changes: the sum over
training instances has been removed, and the regularization contains
an additional factor of $1/N$.  These ensure that the batch gradient
equals the sum of the per-instance gradients, i.e., $\nabla \ell =
\sum_{i=1}^N \nabla \ell_i$, where we use $\nabla \ell_i$ to denote the 
gradient for instance $i$.

At each iteration $m$ of SGD, we randomly select a training instance
$(\inst{\xs}{i},  \inst{\ys}{i})$.  Then compute the new parameter
vector $\theta^{(m)}$ from the old vector $\theta^{(m)}$ by
\begin{equation}
  \label{eq:sgd}
  \theta^{(m)} = \theta^{(m-1)} - \alpha_m \nabla \ell_i(\theta^{(m-1)}),
\end{equation}
where $\alpha_m > 0$ is a step size parameter that controls how far the
parameters move in the direction of the gradient.  If
the step size is too large, then the parameters will swing too far in
the direction of whatever training instance is sampled at
each iteration.  If $\alpha_m$ is too small, then training will proceed
very slowly, to the extent that in extreme cases, the parameters may
appear to have converged numerically when in fact they are far from
the minimum.

We want $\alpha_m$ to decrease as $m$ increases, so that the optimization
algorithm converges to a single answer.  The most common way to do
this is to select a step size schedule of a form like $\alpha_m \sim
1/m$ or $\alpha_m \sim 1/\sqrt{m}$.  
These choices are motivated by the classic convergence results for
stochastic approximation procedures
\cite{robbins-monro,kiefer-wolfowitz}.
However, simply taking $\alpha_m = 1/m$ is usually bad, because then
the first few step sizes are too large.  Instead, a common trick is to use
a schedule like
\begin{equation}
  \label{eq:step-size}
  \alpha_m = \frac{1}{\sigma^2(m_0 + m)},
\end{equation}
where $m_0$ is a free parameter that needs to be set.
A suggestion for setting this parameter, due to Leon Bottou
\cite{bottou:crfweb}, is to sample a
small subset of the training data and run one pass of SGD over the
subset with various fixed step sizes $\alpha$. 
Pick the $\alpha^*$ such that the resulting likelihood on the subset
after one pass is highest, and choose $m_0$ such that $\alpha_0 = \alpha^*$.


Stochastic gradient descent has also gone by the name of backpropagation
in the neural network literature, and many tricks for tuning the method have been developed
over the years \cite{lecun98backprop}. 
Recently, there has been renewed interest in advanced online
optimization methods
\cite{vishwanathan06smd,crammer06online,shalev07pegasos,globerson07exponentiated}, which also update parameters
in an online fashion, but in a more sophisticated way than simple SGD.
\citeauthornum{vishwanathan06smd} was the first application of
stochastic gradient methods to CRFs.

The main disadvantage of stochastic gradient methods is that they do
require tuning, unlike off-the-shelf solvers such as conjugate
gradient and L-BFGS.  Stochastic gradient methods are also not useful
in relational settings in which the training data are not iid, or on
small data sets.  On
appropriate data sets, however, stochastic gradient methods can offer
considerable speedups.

\section{Parallelism}
\label{sec:parallel}

Stochastic gradient descent speeds up the gradient computation by
computing it over fewer instances.  An alternative way to speed up the
gradient computation is to compute the gradient over multiple
instances in parallel.
Because the gradient~\eqref{sm:eqn:lc-lik-derivative} is a sum over
training instances, it is easy to
divide the computation into multiple threads, where each thread
computes the gradient on a subset of training instances.
If the CRF implementation is run on a multicore machine, then the threads will run in
parallel, greatly speeding up the gradient computation.
This is a  characteristic shared by many common machine learning
algorithms, as pointed out by \citeauthornum{chu07map}.

In principle, one could also distribute the gradient computation
across multiple machines, rather than multiple cores of the same
machine, but the overhead involved in transferring large parameter
vectors across the network can be an issue.
A potentially promising way to avoid this is to update the parameter
vectors asynchronously.  An example of this idea is recent
work on incorporating parallel computation into stochastic gradient methods
\cite{langford09slow}.

\section{Approximate Training}
\label{sec:alternative-training}

All of the training methods that we have described so far, including
the stochastic and parallel gradient methods, assume that the
graphical structure of the CRF is tractable, that is, that we can
efficiently compute the partition function $Z(\bx)$ and the marginal
distributions $p(\by_c | \bx)$.  This is the case, for example,
in linear chain and tree-structured CRFs.
Early work on CRFs focused on these cases,
both because of the tractability of inference, and because this choice is
very natural for certain tasks such as sequence labeling tasks in NLP.

But more complex graphs are important in domains such as computer
vision, where grid-structured graphs are natural, and for more global
models of natural language
\cite{sutton04skip,finkel05incorporating,bunescu04cie}.
When the graphical structure is more complex, then the
marginal distributions and the partition function cannot be computed
tractably, and we must resort to approximations.  As described
in Chapter~\ref{chp:inference}, there is a large
literature on approximate inference algorithms.
In the context of CRFs, however, there is a crucial additional
consideration, which is that the approximate inference procedure is
embedded within a larger optimization procedure for selecting the
parameters.

There are two general ways to think about approximate training in CRFs
\cite{sutton06local}: One can either modify the likelihood, or
approximate the marginal distributions directly.  Modifying the
likelihood typically means finding some substitute for $\ell(\theta)$
(such as the BP approximation~\eqref{mdl:eq:l-bethe-primal}), which we will
call a \emph{surrogate likelihood} that is easier to compute but is
still expected to favor good parameter setting. Then the
surrogate likelihood can be optimized using a gradient-based method, in a similar way
to the exact likelihood.  Approximating the marginal distributions
means using a generic inference algorithm to compute an approximation
to the marginals $p(\by_c | \bx)$, substituting the approximate
marginals for the exact marginals in the gradient \eqref{sm:eqn:gradient1}, 
and performing some kind of gradient descent procedure using the
resulting approximate gradients.

Although surrogate likelihood and approximate marginal methods are
obviously closely related, they are distinct.
Usually an surrogate likelihood method directly yields an
approximate marginals method, because just as the derivatives of $\log
Z(\bx)$ give the true marginal distributions, the derivatives of an
approximation to $\log {Z}(\bx)$ can be viewed as an approximation to
the marginal distributions.  These approximate marginals are sometimes termed
\emph{pseudomarginals} \cite{wainwright05estimating}.
However, the reverse direction does not always hold:
for example, there are certain
approximate marginal procedures that provably do not correspond to the derivative
of any likelihood function \cite{sutton06local,sutskever10cd}.

The main advantage of a surrogate likelihood method is that having
an objective function can make it easier to understand the properties
of the method, both to human analysts and to the optimization procedure.
Advanced optimization engines such as conjugate gradient and BFGS
require an objective function in order to operate.
The advantage to the approximate marginals viewpoint, on the other
hand, is that it is more flexible.  It is easy to  incorporate
arbitrary inference algorithms, including tricks such
as early stopping of BP and MCMC.  Also, approximate marginal methods
fit well within a stochastic gradient framework.

There are aspects of the interaction between approximate inference and
parameter estimation that are not completely understood.  For example, \citeauthornum{kulesza08approximate}
present an example of a situation in which the perceptron algorithm
interacts in a pathological fashion with max-product belief
propagation.
Surrogate likelihood methods, by contrast, do
not seem to display this sort of pathology, 
as \citeauthornum{wainwright05estimating} point out for the case of
convex surrogate likelihoods.

To make this discussion more concrete, in the rest of this section, we
will discuss several examples of surrogate likelihood and
approximate marginal methods.  We discuss surrogate likelihood methods
based on pseudolikelihood (Section~\ref{est:sec:pl}) and belief
propagation (Section~\ref{est:sec:bp}) and approximate gradient methods
based on belief propagation (Section~\ref{est:sec:bp}) and MCMC (Section~\ref{est:sec:mcmc}).

\subsection{Pseudolikelihood}
\label{est:sec:pl}


One of the earliest surrogate likelihoods is
the
pseudolikelihood \cite{besag75nonlattice}.
The idea in pseudolikelihood is for the training objective to depend only on 
conditional distributions over single variables.  Because the
normalizing constants for these distributions depend only on single
variables, they can be computed efficiently.  In the context of CRFs,
the pseudolikelihood is
\begin{equation}
\ell_{\PL} (\theta) = \sum_{s \in V} \log p(y_s | \by_{N(s)}, \bx ; \theta)
\end{equation}
Here the summation over $s$ ranges over all output nodes in the graph, and
$\by_{N(s)}$ are the values of the variables $N(s)$ that are neighbors
of $s$.
(As in \eqref{eq:genl-lik}, we do not include the sum over training
instances explicitly.)

Intuitively, one way to understand pseudolikelihood is that it
attempts to match the local conditional distributions $p(y_s | \by_{N(s)}, \bx ; \theta)$ according to
the model  to those of the training data, and because of the
conditional independence assumptions of the model, the local
conditional distributions are sufficient to specify the joint.  (This
is similar to the motivation behind a Gibbs sampler.)

The parameters are estimated by maximizing the pseudolikelihood, i.e.,
the estimates are $\hat{\theta}_{\PL} = \max_{\theta}
\ell_{\PL}(\theta)$.  Typically, the maximization is carried out by a
second order method such as limited-memory BFGS, but in principle parallel computation or stochastic
gradient can be applied to the pseudolikelihood exactly in the same
way as the full likelihood.  Also, regularization can be used just as
with maximum likelihood.

The motivation behind pseudolikelihood is computational efficiency.
The pseudolikelihood can be computed and optimized without needing to
compute $Z(\bx)$ or the marginal distributions.
 Although pseudolikelihood has sometimes proved effective in NLP
\cite{toutanova03pos}, more commonly the performance of
pseudolikelihood is poor \cite{sutton05piecewise}, in an intuitively
analogous way that a Gibbs sampler can mix slowly in sequential models. 
One can obtain better performance by performing a ``blockwise'' version of 
pseudolikelihood in which the local terms involve conditional probabilities of
larger regions in the model.  For example, in a linear-chain CRF, one
could consider a per-edge pseudolikelihood:
\begin{equation}
 \ell_{\EPL}(\theta) = \sum_{t=1}^{T-1} \log p(y_t, y_{t+1} | y_{t-1}, y_{t+2}, \theta)
\end{equation}
(Here we assume that the sequence is padded with dummy labels $y_0$ and $y_{T+1}$
so that the edge cases are correct.)
This blockwise version of pseudolikelihood is a special case of composite likelihood
\cite{lindsay88composite,dillon10composite},
for which there are general theoretical results concerning asymptotic
consistency and normality.
Typically larger blocks lead to better parameter estimates,
both in theory and in practice.

\subsection{Belief Propagation}
\label{est:sec:bp}

The loopy belief propagation algorithm
(Section~\ref{sec:belief-propagation}) can be
used within approximate CRF training.
This can be done within either the surrogate likelihood or the
approximate gradient perspectives.

In the approximate gradient algorithm, at every iteration of training,
we run loopy BP on the training input $\bx$, yielding a set of
approximate marginals $q (\by_c)$ for each clique in the
model.  Then we approximate the true gradient \eqref{sm:eqn:gradient1}
by substituting in the BP marginals.  This results in approximate
partial derivatives
\begin{equation} 
\frac{\partial \tilde{\ell}}{\partial \theta_{pk}} =  
 \sum_{\Psi_c \in C_p} f_{pk} (\xs_c, \ys_c) - \sum_{\Psi_c \in C_p} \sum_{\ys'_c} f_{pk} (\xs_c, \ys'_c) q(\ys'_c). 
\end{equation} 
These can be used to update the current parameter setting as
\begin{equation}
  \label{eq:bp-gradient}
  \theta_{pk}^{(t+1)} = \theta_{pk}^{(t)} + \alpha \frac{\partial \tilde{\ell}}{\partial \theta_{pk}}
\end{equation}
where $\alpha > 0$ is a step size parameter.
The advantages of this setup are that it is extremely simple, 
and is especially useful within an outer stochastic gradient approximation.

More interestingly, however, it is also possible to use loopy BP
within a surrogate likelihood setup.  To do this, we
need to develop 
some surrogate function for the true likelihood \eqref{eq:genl-lik}
which has the property that the gradient of the surrogate likelihood
are exactly the approximate BP gradients \eqref{eq:bp-gradient}.
This may seem like a tall order, but fortunately it is possible using
the Bethe free energy described in Section~\ref{sec:belief-propagation}.

Remember from that section that loopy belief propagation can 
be viewed as an optimization algorithm, namely, one that minimizes the
objective function $\varO_\bethe(q)$ \eqref{bk:eq:betheprimal} over the set of all locally consistent belief vectors,
and that the minimizing value  $\min_q \varO_\bethe(q)$ can be used as an
approximation to the partition function.  Substituting in that approximation
to the true likelihood \eqref{eq:genl-lik} gives us, for a fixed belief vector $q$, 
the approximate likelihood
\begin{multline}
 \ell_\bethe(\theta, q) = \sum_{C_p\in \smCalC} \sum_{\Psi_c \in C_p} \log \Psi_{c} (\xs_c, \ys_c) - \sum_{C_p\in \smCalC} \sum_{\Psi_c \in C_p} q(\by_c) \log \frac{q(\by_c)}{\Psi_c  (\xs_c, \ys_c) } \\ + \sum_{s \in Y} (1 - d_i) q(y_s) \log q(y_s). \label{mdl:eq:l-bethe-primal}
\end{multline}
Then approximate training can be viewed as the optimization problem $\max_\theta \min_q   \ell_\bethe(\theta, q)$.
This is a \emph{saddlepoint problem}, in which we are maximizing with respect to one variable (to find the best parameters)
and minimizing with respect to another (to solve the approximate inference problem).
One approach to solve saddlepoint problems is coordinate ascent, that is,
to alternately minimize $\lBethe$ with respect to $q$ for fixed $\theta$
and take a gradient step to partially maximize
$\lBethe$  with respect to $\theta$ for fixed $b$.
The first step (minimizing with respect to $q$) is just running the loopy BP algorithm.
The key point is that for the second step (maximizing with respect to $\theta$),
the partial derivatives of \eqref{mdl:eq:l-bethe-primal} with respect
to a weight $\theta_k$ is exactly \eqref{eq:bp-gradient}, as desired.

Alternatively, there is a different surrogate likelihood that can also be used.
This is
\begin{equation}
\lapp(\theta; q) = \log \left\lbrack \frac{\prod_{C_p\in \smCalC} \prod_{\Psi_c \in C_p}  q (\by_c)}{\prod_{s \in Y} q (y_s) ^{d_s-1}} \right\rbrack,
\label{est:eq:l-bethe-dual}
\end{equation}
In other words, instead of the true joint likelihood, we use the 
product over each clique's approximate belief, dividing by the 
node beliefs to avoid overcounting.
The nice thing about this is that it is
a direct generalisation of the true likelihood for tree-structured
models, as can be seen by comparing \eqref{est:eq:l-bethe-dual} with
\eqref{bk:eq:treejoint}.  This surrogate likelihood can be justified
using a dual version of Bethe energy that we have presented here \cite{minka01ep-energy,minka05divergence}.
When BP has converged, for the resulting belief vector $q$, it can be shown that
$\ell_\bethe(\theta, q) = \lapp(\theta, q)$.  This equivalence does not hold
in general for arbitrary values of $q$, e.g., if BP has not converged.

Another surrogate likelihood method that is related to BP is
the \emph{piecewise} estimator \citep{sutton08piecewise},
in which the factors of the model are partitioned 
into tractable subgraphs that are trained independently.
This idea can work surprisingly well (better than pseudolikelihood) if
the local features are sufficiently informative.
\citeauthornum{sutton06local} discuss the close relationship between
piecewise training and early stopping of belief propagation.

\subsection{Markov Chain Monte Carlo}
\label{est:sec:mcmc}

Markov Chain Monte Carlo (MCMC) inference methods
(Section~\ref{sec:mcmc})
can be used within CRF training by setting up a Markov chain whose
stationary distribution is $p(\by | \bx; \theta)$, running the chain
for a number of iterations, and using the resulting approximate
marginals $\hat{p}(\by | \bx; \theta)$ to approximate the true
marginals in the gradient \eqref{sm:eqn:gradient1}.

In practice, however, MCMC methods are not commonly used in the
context of CRFs.
There are two main reasons for this.  First, MCMC methods typically
require many iterations to reach convergence, and as we have
emphasized, inference needs to be run for many different
parameter settings over the course of training.
Second, many MCMC methods, such as Metropolis-Hastings, require
computing a ratio of normalising constants $Z_{\theta_1}(\bx) /
Z_{\theta_2}(\bx)$ for two different parameters settings $\theta_1$
and $\theta_2$.  This presents a severe difficulty for models in which
computing $Z_\theta(\bx)$ is intractable.

One possibility to overcome these difficulties is contrastive divergence (CD)
\cite{hinton02cd}, in which the true marginals $p(y_c | \bx)$ in
\eqref{sm:eqn:gradient1} are approximated by running an MCMC method
for only a few iterations, where the initial state of the Markov chain
(which is just an assignment to $\by$) is set to be the value of $\by$
in the training data.
CD has been mostly applied to latent variable models such as
restricted Boltzmann machines, it can also be applied to CRFs.
We are unaware  of much work in this direction.

Another possibility is a more recent method called SampleRank
\cite{wick09samplerank}, whose objective is that the learned
parameters score pairs of {\bf y}s such that their sorted ranking
obeys a given supervised ranking (which is often specified in terms of
a fixed scoring function on {\bf y} that compares to true target
values of {\bf y}).  Approximate gradients may be calculated from
pairs of successive states of the MCMC sampler.  Like CD, SampleRank
learns very quickly because it performs useful parameter updates on
many individual MCMC steps.  Experiments have shown the structured
classification accuracy from SampleRank to be substantially higher
than CD \cite{wick09samplerank}.

The discussion above concerns MCMC methods within an approximate
gradient framework.
In contrast, it is very difficult to use an MCMC inference method
within an surrogate likelihood framework, because it is notoriously
difficult to obtain a good approximation to $\log Z(\bx)$ given
samples from an MCMC method.

\begin{table}
\hspace{-0.75in}
  \begin{tabular}{l||rrrrr||r}
 Task & Parameters & Predicates & \# Sequences & \# Positions & Labels & Time (s) \\
\hline 
NP chunking & 248471 & 116731 & 8936 & 211727 & 3 & 958s \\
NER & 187540 & 119265 & 946 & 204567 & 9 & 4866s\\
POS tagging & 509951 & 127764 & 38219 & 912344 & 45 &  325500s \\
\hline
  \end{tabular}
  \caption{Scale of typical CRF applications in natural language processing}
  \label{tbl:crf:application}
\end{table}

\section{Implementation Concerns}
\label{sec:times}

To make the discussion of efficient training methods more concrete,
here we give some examples of data sets from NLP in which CRFs have been successful.
The idea is to give a sense of the scales of problem to which CRFs have been applied,
and of typical values of the number of the numbers of features and of training times.

We describe three example tasks to which CRFs have been applied.  The first example task is noun-phrase (NP) chunking \cite{conll2000}, in which the problem is to find 
base noun phrases in text, such as the phrases ``He'' and ``the current account deficit''
in the sentence \emph{He reckons the current account deficit will narrow}.  
The second task is named identity recognition (NER) \cite{conll2003},
The final task is part-of-speech tagging (POS), that is, labelling each word in a sentence with its part of speech. 
The NP chunking and POS data sets are derived from the WSJ Penn Treebank \cite{marcus93building},
while the NER data set consists of newswire articles from Reuters.

We will not go into detail about the features that we use, but they include the identity of the current and previous word,
prefixes and suffixes, and (for the named-entity and chunking tasks) automatically generated part of speech tags and lists of common
places and person names.
We do not claim that the feature sets that we have used are optimal for these tasks,
but still they should be useful for getting a sense of scale.

For each of these data sets, Table~\ref{tbl:crf:application} shows (a)
the number of parameters in the trained CRF model, (b) the size of the
training set, in terms of the total number of sequences and number of
words, (c) the number of possible labels for each sequence position,
and (d) the training time.
The training times range from minutes in the best case to days in the worst case.
As can be expected from our previous discussion, the factor 
that seems to most influence training time is the number of labels.

Obviously the exact training time will depend heavily on 
details of the implementation and hardware.
For the examples in Table~\ref{tbl:crf:application}, we use the MALLET toolkit
on machines with a 2.4 GHz Intel Xeon CPU, optimizing the likelihood
using batch L-BFGS without using multithreaded or stochastic gradient
training.

\chapter{Related Work and Future Directions}
\label{chp:related}

In this section, we briefly place CRFs in the context of related lines of research, especially that of \emph{structured prediction},
a general research area which is concerned with extending classification methods to complex objects.
We also describe relationships both to neural networks and to a simpler sequence model called maximum entropy Markov models (MEMMs).
Finally, we outline a few open areas for future work.

\section{Related Work}

\subsection{Structured Prediction}

Conditional random fields provide one method for extending the ideas
behind classification to the prediction of more complex objects such
as sequences and trees.
This general area of research is called \emph{structured prediction}.
Essentially, logistic regression is to a CRF as classification is to
structured prediction. 
Examples of the types of structured outputs that are considered include parse trees of natural language
sentences \cite{taskar04mmcfg,finkel08efficient}, alignments between sentences in different languages
\cite{taskar05alignment}, and route plans in mobile robotics  \cite{ratliff06planning}.
 Detailed information about structured
prediction methods is available in a recent collection of research
papers \citep{bakir07structured}.

Structured prediction methods are essentially a combination of
classification and graphical modeling, combining the ability to
compactly model multivariate data with the ability to perform
prediction using large sets of input features.  The idea is, for an input $\bx$,
to define a discriminant function $F_{\bx} (\by)$, and predict $\by^{*} = \arg \max_{\by} F_{\bx} (\by)$.  This function
factorizes according to a set of local factors, just as in graphical
models.  But as in classification, each local factor is modeled a linear function of $\bx$, although perhaps in some induced high-dimensional space.  To understand
the benefits of this approach,
consider a hidden Markov model (Section~\ref{sec:bk:sequence}) and a set of per-position classifiers, both with fixed parameters.  In principle, the per-position classifiers predict
an output $y_{s}$ given all of $\bx_{0} \ldots \bx_{T}$.\footnote{To be fair, in practice the classifier for $y_{s}$ would probably depend only on a sliding window around $\bx_{s}$, rather than all of $\bx$.}  In the HMM, on the other hand, to predict $y_{s}$ it is statistically sufficient to know only the local input $\bx_{s}$, the previous forward message $p(y_{s-1}, \bx_{0} \ldots \bx_{s-1})$, and the backward message $p(\bx_{s+1} \ldots \bx_{T} | y_{s})$.  So the forward and backward messages
serve as a summary of the
rest of the input, a summary that is generally non-linear in the
observed features.

In principle, the same effect could be achieved using a per-position classifier
if it were possible to define an extremely flexible set of nonlinear features that
depend on the entire input sequence.
But as we have seen the size of the input vector is extremely large.  For
example, in part-of-speech tagging, each vector $\bx_{s}$ may have
tens of thousands of components, so a classifier based on all of $\bx$
would have many parameters.  But using only $\bx_{s}$ to
predict $y_{s}$ is also bad, because information from
neighboring feature vectors is also useful in making predictions.
Essentially the effect of a structured prediction method is that
a confident prediction about
one variable is able to influence nearby, possibly less
confident predictions.

Several types of structured prediction algorithms have been studied.
All such algorithms assume that the discriminant function $F_{\bx} (\by)$ over labels can be written as a sum of local functions $F_{\bx}(\by) = \sum_{a} f_{a} (\by_{a}, \bx, \theta)$.
The task is to estimate the real-valued parameter vector $\theta$
given a training set  $\Data = \{ \inst{\xs}{i}, 
\inst{\ys}{i} \}_{i=1}^N$.
The methods differ in how the parameters are selected.

Alternative structured prediction methods are based on maximizing over assignments rather than marginalizing.
Perhaps the most popular of these methods has been
\emph{maximum-margin} methods that are so successful for univariate
classification.   Maximum margin methods have been generalized to the structured case
\cite{altun03hmsvm,taskar03m3ns}.  Both batch and online algorithms
have been developed to
maximize this objective function.  
The perceptron update can also be generalized 
to structured models
\cite{collins02discriminative}.  The resulting algorithm is particularly appealing because it is little more difficult to implement than the algorithm for selecting $\by^{*}$.
The online perceptron update can also be made margin-aware,
yielding the MIRA algorithm \cite{crammer03ultraconservative},
which may perform better than the perceptron update.

Another class of methods are search-based methods \cite{daume05laso,daume09searn} in which a
heuristic search procedure over outputs is assumed, and learns a
classifier that predicts the next step in the search.  This has the
advantage of fitting in nicely to many problems that are complex
enough to require performing search.  It is also able to incorporate
arbitrary loss functions over predictions.

A general advantage of all of these
maximization-based methods is that they do not require summation over all configurations for the partition function or for marginal distributions.  There are certain combinatorial
problems, such as matching and network flow problems, in which finding
an optimal configuration is tractable, but summing over configurations
is not (for an example of applying max-margin methods in such situations, see \citeauthornum{taskar05alignment}).  For more complex problems, neither summation nor maximization is tractable, so this advantage is perhaps not as significant.
Another advantage of these methods is that kernels can be naturally
incorporated, in an analogous way as in support vector machines.

Finally, \citeauthornum{lecun06energy} generalizes many prediction methods,
including the ones listed above,
 under the rubric of \emph{energy-based} methods, and presents
interesting historical information about their use.  They
advocate changing the loss function to avoid probabilities
altogether.

Perhaps the main advantage of probabilistic methods is that they can incorporate latent variables in a natural way,
by marginalization.  This can be useful, for example, in collective
classification methods \cite{taskar02rmn}.
For examples of structured models with latent variables, see
\citeauthornum{quattoni05conditional} and \citeauthornum{mccallum05stringedit}.
 A particularly powerful example of this is provided by Bayesian methods, in which the model parameters themselves are integrated out (Section~\ref{sec:bayesian}).

The differences between the various structured prediction methods are
not well understood.  
To date, there has been little careful comparison of these, 
especially CRFs and max-margin approaches, across  
different structures and domains, although see \citeauthornum{keerthi07crf} for some experiments in this regard.\footnote{An earlier study \cite{nguyen07comparison} appears to have been flawed.  See  \citeauthornum{keerthi07crf}  for discussion.}
We take the view that the similarities between various structured prediction methods are more important than the differences. 
Careful selection of features has more effect on performance than the choice of structured prediction algorithm.

\subsection{Neural Networks}

There are close relationships between neural
networks and conditional random fields, in that both can be viewed as
discriminatively trained probabilistic models.  
Neural networks are perhaps best known for their use in classification, but they can also be used to predict multiple outputs,
for example, by using a shared latent representation \cite{caruana97multitask}, or
by modelling dependencies between outputs directly \cite{lecun98gradient}.
Although neural networks are typically trained using stochastic
gradient descent (Section~\ref{est:sec:sg}), in principle they can
be trained using any of the other methods used for CRFs.
The main difference between them is that neural networks
represent the dependence between output variables using a shared
latent representation, while structured methods learn these
dependences as direct functions of the output variables.

Because of this, it is easy to make the mistake of thinking that CRFs are convex and
neural networks are not.  This is incorrect.  A neural network without
a hidden layer is a linear classifier that can be trained efficiently
in a number of ways, while a CRF with latent variables has a complex
non-convex likelihood (Section~\ref{sec:mdl:crf-general}).
The correct way of thinking is: In fully observed models, the
likelihood is convex; in latent variable models it is not.

So the main new insight of structured prediction models
compared to neural networks is: If you add connections among the nodes in the
output layer, and if you have a good set of features, then sometimes you don't need a hidden layer to
get good performance.  If you can afford to leave out the hidden, then in practice you always want to do so, because
this avoids all of the problems with local minima.  For harder
problems, however, one might expect that even after modeling output
structure, incorporating hidden state will still yield additional
benefit.  Once hidden state is introduced into the model, whether it
be a neural network or a structured model, it seems to be inevitable
(at least given our current understanding of machine learning) that
convexity will be lost.

\begin{figure}
  \centering
  \resizebox{3in}{!}{\includegraphics{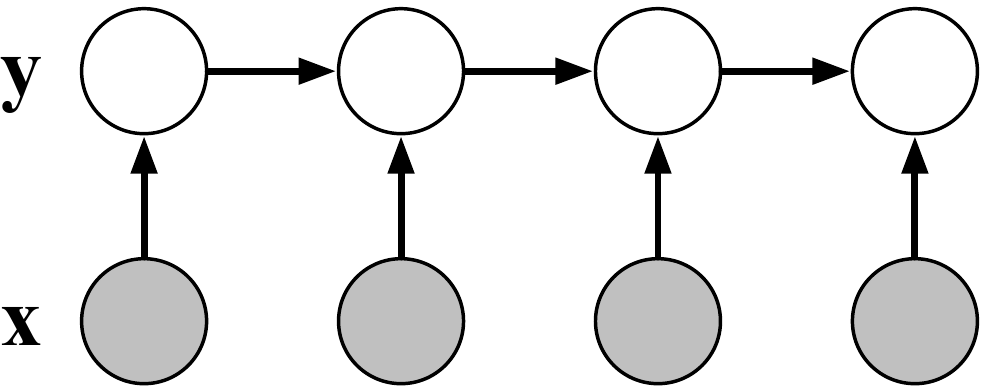}}
  \caption{Graphical model of a maximum entropy Markov model \cite{mccallum00memm}.}
  \label{fig:memm}
\end{figure}
\subsection{MEMMs, Directed Models, and Label Bias}

Linear-chain CRFs were originally introduced as an improvement to the
\emph{maximum-entropy Markov model} (MEMM) \citep{mccallum00memm},
which is essentially a Markov model in which the transition
probabilities are given by logistic regression.  Formally,
an MEMM is
\begin{align}
  \label{eq:memm}
  p_{\MEMM}(\by | \bx) &= \prod_{t=1}^T p(y_t | y_{t-1}, \bx) \\
p(y_t | y_{t-1}, \bx) &= \frac{1}{Z_t(y_{t-1}, \bx)} \exp\left\{ \sum_{k=1}^K  \theta_k f_k (y_t, y_{t-1}, \bx_t) \right\} \\
Z_t(y_{t-1}, \bx) &= \sum_{y'}  \exp\left\{ \sum_{k=1}^K  \theta_k f_k (y', y_{t-1}, \bx_t) \right\} 
\end{align}
A similar idea can be extended to general directed graphs, in which the distribution $p(\by | \bx)$ is expressed by a Bayesian network
in which each CPT  is a logistic regression models with input $\bx$
 \cite{rosenberg07skip-memm}.

 In the linear-chain case, notice that the MEMM works out to have the
 same form as the linear-chain CRF \eqref{sm:eqn:lc-crf-likelihood}
 with the exception that  in a CRF $Z(\bx)$ is a sum over sequences,
whereas in a MEMM the analogous term is $\prod_{t=1}^T Z_t(y_{t-1},
 \bx)$.  This difference has important consequences.  Unlike in a
 CRFs, maximum likelihood training of MEMMs does not require
 performing inference, because $Z_t$ is just a simple sum over the
 labels at a single position, rather than a sum over labels of an
 entire sequence.  This is an example of the general
 phenomenon that training of directed models is less computationally
 demanding than undirected models.

There are theoretical difficulties with the MEMM model, however.
MEMMs can exhibit the problems of label bias \citep{lafferty01crf} and observation bias \citep{klein02conditional}. 
Originally, the label bias problem was described from an algorithmic perspective.
Consider the backward recursion \eqref{sm:eqn:hmm-bwd}.  In the case of an MEMM, this amounts to
\begin{equation} 
\beta_t (i) = \sum_{j \in S} p(y_{t+1} = j | y_t = i, x_{t+1}) \beta_{t+1}(j).\label{eq:memm-beta}
\end{equation} 
Unfortunately, this sum is always 1, regardless of the value of the current label $i$.  
To see this, assume for the sake of induction that $\beta_{t+1}(j) =
1$ for all $j$.  Then it is clear that the sum over $j$ in
\eqref{eq:memm-beta} collapses, and $\beta_t(i) = 1$.
What this means is that the future observations provide no information about the current state,
which seems to lose a major advantage of sequence modelling.

Perhaps a more intuitive way to understand label bias is from
the perspective of graphical models.
Consider the graphical model of an MEMM, shown in Figure~\ref{fig:memm}.
By looking at the v-structures in the graph, we can read off the following independence assumptions:
at all time steps $t$, the label $y_t$ is marginally independent of the future observations 
$\bx_{t+1}, \bx_{t+2},$ etc.
This independence assumption is usually strongly violated in
sequence modeling, which explains why CRFs can have better performance
than MEMMs.  Also, this independence relation explains why $\beta_t (i)$ should always be 1.
(In general, this correspondence between graph structure and inference
algorithms is one of main conceptual advantages of graphical modelling.)
To summarize this discussion, \emph{label bias is simply a consequence of explaining away}.

There is a caveat here:
We can always copy information from previous and future time steps
into the feature vector $\bx_t$, and this is common in practice.  (The only constraint is that if we
have too many features, then overfitting we become an issue.)  This
has the effect of adding arcs between (for example) $\bx_{t+1}$. This explains why the performance gap
between MEMMs and CRFs is not always as large as might be expected.

Finally, one might try a different way to combine the advantages of
conditional training and directed models.  One can imagine defining a
directed model $p(\by, \bx)$, perhaps a generative model, and then
training it by optimizing the resulting conditional likelihood
$p(\by|\bx)$.  In fact, this procedure has long been done in the
speech community, where it is called maximum mutual information
training.  However, this does not have strong computational benefits
over CRFs.  The reason is that
computing the conditional likelihood $p(\by|\bx)$ 
requires computing the marginal probability $p(\xs)$, which
plays the same role as $Z(\xs)$ in the CRF likelihood.  In fact,
training is more complex in a directed model, because the model
parameters are constrained to be probabilities---constraints which can actually
make the optimization problem more difficult. 

\section{Frontier Areas}
 
Finally, we describe a few open research areas that related to CRFs.
In all of the cases below, the research question is a special case of
a larger question for general graphical models, but there are special
additional considerations in conditional models that make the problem
more difficult.

\subsection{Bayesian CRFs}
\label{sec:bayesian}

Because of the large number of parameters in typical applications of CRFs, the models can be prone to overfitting.
The standard way to control this is using regularization, as described in Section~\ref{sec:lc-crf-estimation}.
One way that we motivated this procedure is as an approximation to a fully Bayesian procedure.
That is, instead of predicting the labels of a testing instance $\bx$ as
$\by^* = \max_{\by} p(\by | \bx; \hat{\theta})$, where $\hat{\theta}$ is a single parameter estimate, in a Bayesian method we would use the predictive distribution
$\by^* = \max_{\by} \int p(\by | \bx; \theta) p(\theta) \prod_{i=1}^N p(\inst{\by}{i} | \inst{\bx}{i}, \theta) d\theta$.
This integral over $\theta$ needs to be approximated, for example, by MCMC.

In general, it is difficult to formulate efficient Bayesian methods for undirected models; see \cite{murray06mcmc,murray:thesis} for some of the few examples in this regard.
A few papers have specially considered approximate inference algorithms for Bayesian CRFs \cite{qi05bcrf,welling06bayesian}, but while these methods are interesting, they do not seem to be useful at the scale of current CRF applications (e.g., those in Table~\ref{tbl:crf:application}).
Even for linear chain models, Bayesian methods are not commonly in use for CRFs, primarily due to the computational demands.
If all we want is the benefits of model averaging, one may question whether simpler ensemble learning techniques, such as bagging, would give the same benefit.
However, the Bayesian perspective does have other potential benefits, particularly when more complex, hierarchical priors are considered.

\subsection{Semi-supervised CRFs}

One practical difficulty in applying CRFs is that training requires
obtaining true labels for potentially many sequences.  This can be
expensive because it is more time consuming for a human labeller to
provide labels for sequence labelling than for simple classification.
For this reason, it would be very useful to have techniques that can
obtain good accuracy given only a small amount of labeled data.  

One strategy for achieving this goal is \emph{semi-supervised learning},
in which in addition to some fully-labelled data $\{ (\inst{\bx}{i}, \inst{\by}{i}) \}_{i=1}^N$, the data set is assumed to contain
a large number of unlabelled instances $\{ \inst{\bx}{j} \}_{j=1}^M$, for which we observe only the inputs.
However, unlike in
generative models, it is less obvious how to incorporate unlabelled
data into a conditional criterion, because the unlabelled data is a
sample from the distribution $p(\bx)$, which in principle need have no
relationship to the CRF $p(\by|\bx)$.
In order to deal with this, several different types of regularization terms have been proposed that take the unlabelled data into account,
including entropy regularization \cite{grandvalet04semi,jiao06semi},
generalized expectation criteria \cite{mann08generalized}, posterior
regularization \cite{ganchev09posterior,graca09posterior}, and measurement-based learning
\cite{liang09measurements}.

\subsection{Structure Learning in CRFs}

All of the methods described in this tutorial assume that the
structure of the model has been decided in advance.  It is natural to
ask if we can learn the structure of the model as well.  As in
graphical models more generally, this is a difficult problem.  In
fact, \citeauthornum{bradley10crf} point out an interesting
complication that is specific to conditional models.  Typically,
maximum likelihood structure learning can be performed efficiently if
the model is restricted to be tree-structured, using the well-known
Chow-Liu algorithm.  The analogous algorithm in the conditional case
is more difficult, however, because it requires estimating marginal
distributions of the form $p(y_{u}, y_{v} | \bx_{1:N})$, that is, we
need to estimate the effects of the entire input on every pair of
variables.  It is difficult to estimate these distributions
efficiently without knowing the structure of the model to begin with.

\section*{Acknowledgments} 
 
We thank Francine Chen, Benson Limketkai, Gregory Druck, Kedar
Bellare, and Ray Mooney for useful comments 
on earlier versions of this tutorial.  A previous version of this
tutorial has appeared in \citeauthornum{sutton07introduction},
and as part of Charles Sutton's doctoral dissertation \cite{sutton:thesis}.
 

\bibliographystyle{plainnat}
\bibliography{database}

\end{document}